\documentclass[journal,twoside,web]{ieeecolor}
\usepackage{jsen}
\usepackage{cite}
\usepackage{amsmath,amssymb,amsfonts}
\usepackage{algorithmic}
\usepackage{hyperref, graphicx}
\usepackage{textcomp}
\usepackage{wrapfig}
\usepackage{tabu,float,textcomp}
\usepackage{color,booktabs,relsize}
\usepackage{adjustbox}
\usepackage{multirow,dcolumn,hhline}
\usepackage[subrefformat=parens,labelformat=parens,caption=false]{subfig}

\def\BibTeX{{\rm B\kern-.05em{\sc i\kern-.025em b}\kern-.08em
    T\kern-.1667em\lower.7ex\hbox{E}\kern-.125emX}}
\definecolor{abstractbg}{rgb}{0.89804,0.94510,0.83137}
\begin{document}
    \title{Real-Time Damage Detection in Fiber Lifting Ropes Using Lightweight Convolutional Neural Networks}
\author{Tuomas Jalonen, \IEEEmembership{Graduate Student Member, IEEE},
        Mohammad Al-Sa'd, \IEEEmembership{Senior Member, IEEE},
        Roope Mellanen,
        Serkan Kiranyaz, \IEEEmembership{Senior Member, IEEE},
        and Moncef Gabbouj, \IEEEmembership{Fellow, IEEE}
\thanks{This work was funded by Konecranes Plc as part of Business Finland and DIMECC Intelligent Industrial Data Program. \textit{(Corresponding author: Tuomas Jalonen.)}}
\thanks{Tuomas Jalonen and Moncef Gabbouj are with the Faculty of Information Technology and Communication Sciences, Tampere University, 33720 Tampere, Finland (e-mail: \href{mailto:tuomas.jalonen@tuni.fi}{tuomas.jalonen@tuni.fi}; \href{mailto:moncef.gabbouj@tuni.fi}{moncef.gabbouj@tuni.fi}).}
\thanks{Mohammad Al-Sa'd is with the Faculty of Medicine, University of Helsinki, 00014 Helsinki, Finland (e-mail: \href{mailto:mohammad.al-sad@helsinki.fi}{mohammad.al-sad@helsinki.fi}) and the Faculty of Information Technology and Communication Sciences, Tampere University, 33720 Tampere, Finland \href{mailto:mohammad.al-sad@tuni.fi}{mohammad.al-sad@tuni.fi}).
}
\thanks{Roope Mellanen is with Konecranes Plc, 05830 Hyvinkää, Finland (e-mail: \href{mailto:roope.mellanen@konecranes.com}{roope.mellanen@konecranes.com}).}
\thanks{Serkan Kiranyaz is with the Department of Electrical Engineering, Qatar University, 2713 Doha, Qatar (e-mail: \href{mailto:mkiranyaz@qu.edu.qa}{mkiranyaz@qu.edu.qa}).}
}

\IEEEtitleabstractindextext{%
\fcolorbox{abstractbg}{abstractbg}{%
\begin{minipage}{\textwidth}%
\begin{wrapfigure}[20]{r}{3.4in}%
    \includegraphics[width=3.4in]{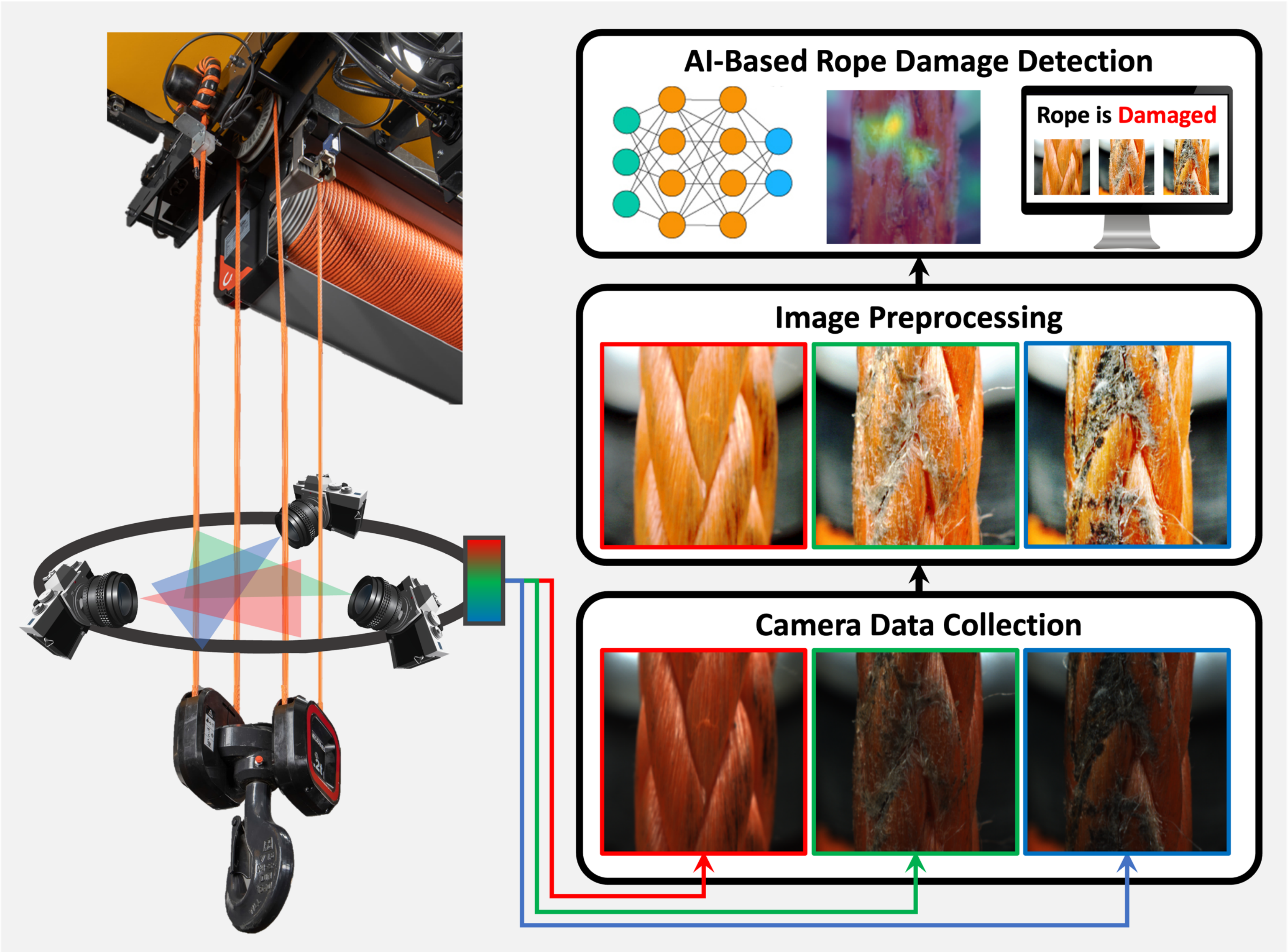}%
\end{wrapfigure}%
\begin{abstract}
The health and safety hazards posed by worn crane lifting ropes mandate periodic inspection for damage. This task is time-consuming, prone to human error, halts operation, and may result in the premature disposal of ropes. Therefore, we propose using efficient deep learning and computer vision methods to automate the process of detecting damaged ropes.
Specifically, we present a vision-based system for detecting damage in synthetic fiber rope images using lightweight convolutional neural networks. We develop a camera-based apparatus to photograph the lifting rope's surface, while in operation, and capture the progressive wear-and-tear as well as the more significant degradation in the rope's health state. Experts from Konecranes annotate the collected images in accordance with the rope's condition; normal or damaged. Then, we pre-process the images, systematically design a deep learning model, evaluate its detection and prediction performance, analyze its computational complexity, and compare it with various other models.
Experimental results show the proposed model outperforms other similar techniques with 96.5\% accuracy, 94.8\% precision, 98.3\% recall, 96.5\% F1-score, and 99.3\% AUC. Besides, they demonstrate the model's real-time operation, low memory footprint, robustness to various environmental and operational conditions, and adequacy for deployment in industrial applications such as lifting, mooring, towing, climbing, and sailing.
\end{abstract}

\end{minipage}}}

\maketitle

\section{Introduction}

\begin{figure}[!t]
\centerline{\includegraphics[width=\columnwidth]{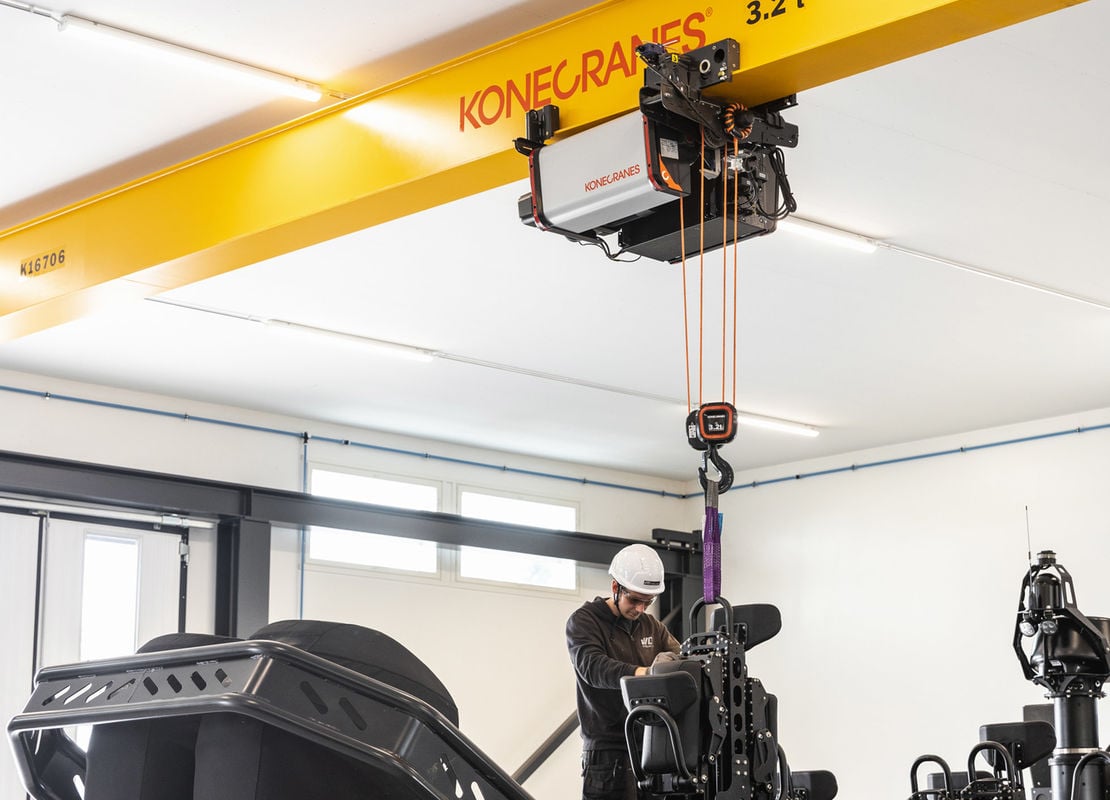}}
\caption{The fiber rope crane used in this work. The photo is published with permission from Konecranes \cite{Crane_manual}.}
\label{figure:crane}
\end{figure}

\IEEEPARstart{R}{ecent} advances in artificial intelligence and computer vision are constantly increasing productivity and safety in manufacturing and logistics \cite{XU2021107530,Jalonen2021}. Nonetheless, lifting heavy payloads is still a major health and safety hazard in many environments related to transportation \cite{liu2023optimal,10082943,lee2020causes}. For example, smaller cranes, like the one shown in Fig. \ref{figure:crane}, can move objects weighing several metric tons. Not to mention larger ones with the capacity for hundreds of tons. Therefore, it is paramount to inspect their ropes for damage to prevent serious accidents, injuries, and additional costs \cite{li2020bi,Konecrane_crane}.
More specifically, lifting ropes are major points of failure that require periodic inspection and replacement \cite{Mupende_Lukasch_Tonnier_2020}. However, manual inspection procedures are labor intensive, time-consuming, subjective, and often require halting the production process \cite{li2020bi,8706985,ISO_standrad,Schmieder_Golder_2020,falconer2017preliminary}.
Therefore, we propose a visual real-time damage detection system for synthetic fiber lifting ropes based on deep learning and computer vision techniques.

The main contributions of this paper are:
\begin{itemize}
    \item Creating an imaging apparatus that photographs crane ropes for visual damage inspection.
    \item Developing the first fiber lifting rope image dataset\footnote{Please contact Roope Mellanen at \href{mailto:roope.mellanen@konecranes.com}{roope.mellanen@konecranes.com} for data inquiry.}.
    \item Designing an AI-based real-time industrial solution for detecting damage in fiber rope images\footnote{The implementation is available at \url{https://github.com/TuomasJalonen/rope-damage-sensors}.}.
\end{itemize}

Lifting ropes are commonly manufactured from steel wires or lately using synthetic fibers such as polyethylene \cite{MIC202031}. Synthetic lifting ropes have many benefits over traditional steel wire ropes. For example, they demonstrate higher corrosion resistance, do not require greasing, and are easier to install \cite{Foster2002}. Moreover, despite their higher purchase price, synthetic fiber ropes are lightweight which allows for utilizing smaller cranes; leading to cost reductions \cite{oland2017condition}. In addition to lifting, fiber ropes have countless applications in the industry, including towing, mooring, cable pulling, sailing, and climbing. However, synthetic ropes, just like steel wires, do suffer from wear and tear and get damaged over time. 
They are primarily subjected to two types of fatigue: tension–tension fatigue due to fluctuating tensile loads, and bending-over sheaves fatigue resulting from repetitive bending \cite{rani2024survey}. Other factors that affect their fatigue lifespan include strand cuts, abrasion, melting, compression damage, pulled strands, and inconsistent diameter.
In contrast to steel wires which tend to break from the inside \cite{schlanbusch2017condition}, synthetic rope damages manifest on the rope's surface and can be visually inspected by an expert \cite{oland2017condition}.
Currently, monitoring the condition of synthetic fiber ropes is performed manually by inspectors following the ISO-9554 standard \cite{ISO_standrad,Schmieder_Golder_2020}. Although it is the standard practice, this procedure is cumbersome, discontinuous in time, interrupts operation, and may result in the premature disposal of ropes \cite{falconer2017preliminary}. Thus, the full benefits of scheduling optimization research in e.g., container terminals \cite{li2020integrated, cahyono2021simultaneous} and cranes \cite{zheng2018two, cahyono2019discrete} may be unreachable with the present method.
Therefore, automatic damage detection by leveraging the recent advancements in computer vision, image processing, and deep learning techniques is needed \cite{oland2017condition}. 
On the one hand, image processing, and related feature extraction methods utilize expert knowledge and attempt to characterize damage in rope images similar to the ones identified by expert inspectors \cite{app9132771}. These techniques generally perform well in a controlled environment, but they poorly integrate the varying conditions and operations found in a real-life setting e.g., noise, lighting conditions, oil residue, and dust \cite{HUANG2020107843,Zhou2021}.
On the other hand, deep learning tools discard the notion of hand-crafted features by learning abstractions that maximize the detection of damaged ropes. In fact, they yield discriminatory features without predisposition to the standard markings and can accommodate a wider range of environmental and/or operational conditions \cite{HUANG2020107843}. For safety-critical applications, the security versus accuracy trade-off of neural networks can be optimized \cite{10070872}. Therefore, deep learning techniques are more suited to detect damage in synthetic fiber rope images compared to engineering-based feature extraction methods. The possible non-destructive methods are further illustrated in Table \ref{table:method_comparison}. We deduced from the table that using optical cameras is a suitable option as it balances the cost of the apparatus and imaging accuracy.
\begin{table}[!t]
\centering
\caption{Comparison of non-destructive methods suitable for damage detection in fiber lifting ropes. Adapted from \cite{app9132771, santur2022new}.}
\begin{adjustbox}{max width=\columnwidth}
\begin{tabu}{ccc}
\toprule
\textbf{Method} & \textbf{Advantages} & \textbf{Disadvantages}
\\\midrule
Manual & Explainable, accurate & Labor-intensive, halts operation
\\\midrule
X-ray & Detects hidden defects & Safety issues
\\\midrule
Ultrasound Guided Wave & Detects hidden defects, long distances & Low resistance to noise
\\\midrule
Acoustic Emission & Detects defect initiation and growth & High cost
\\\midrule
Optical camera & Accurate, low cost & Only surface defects
\\\midrule
Laser scanner & Very accurate & High cost
\\\midrule
Tension meter & Accurate & High cost
\\\bottomrule
\end{tabu}
\end{adjustbox}
\label{table:method_comparison}
\end{table}

The construct of damage indicators in fiber ropes was first articulated in \cite{falconer2017preliminary} where changes in the rope's width and length were found to be important. This particular finding was verified in \cite{FALCONER2020102248} using computer vision and thermal imaging; however, explicit identification for damaged ropes was not performed. In fact, detecting damage in synthetic fiber rope images has received less attention in the literature compared to steel wire cables.
Fortunately, these detection techniques are suitable for adoption due to the similarity between the two problems; they both deal with detecting damaged yarns or strands in rope images.
For instance, the health condition of balancing tail ropes was monitored in \cite{app8081346} using a convolutional neural network (CNN). The rope image was captured and then fed to the CNN model to classify its health state as either normal or if it suffers from one out of eight common damage types. Also, laser cameras could be used to collect data for CNNs as shown in \cite{santur2022new}.
Moreover, a CNN-based approach was designed in \cite{Zhou2019} to detect surface defects in steel wire rope images. The model classified the acquired images into normal, broken, or damaged, and achieved a 99.7\% overall accuracy. The same problem was tackled in \cite{ZHOU2019106954} using support vector machines trained with texture-based hand-crafted features. The proposed system achieved a 93.3\% classification accuracy and it was further improved in \cite{8972428} to reach 95.9\%. Nonetheless, the limited sample size and the reliance on hand-crafted features hampered robustness in noisy environments. This was evidenced in \cite{s20226612} which showed that the model accuracy drops to 80.5\% when training/testing with a different dataset. Moreover, the utility of CNNs combined with image processing techniques was shown to increase the accuracy of the model in \cite{ZHOU2019106954} from 93.3\% to 95.5\% \cite{Zhou2021}. This has motivated us to design a CNN-based solution for detecting damage in synthetic fiber rope images.
However, our proposed solution will be developed to have both high performance and low computational requirements, allowing for easy integration into industrial systems and efficient deployment \cite{Verhelst2017, 9447019}.

The rest of this paper is organized as follows:
section \ref{sec:methodology} describes our methodology for building the experimental setup, collecting data, designing the damage detection system, and evaluating its performance. Afterwards, we present and discuss the system's performance and compare it to various other models in section \ref{sec:results}. Finally, section \ref{sec:conclusions} concludes the paper and suggests topics for future research.
\section{Methodology} \label{sec:methodology}
The proposed fiber rope damage detection system is overviewed in Fig. \ref{fig:system_model} and consists of the following stages:
\begin{figure*}[!t]
\centerline{\includegraphics[width=\textwidth]{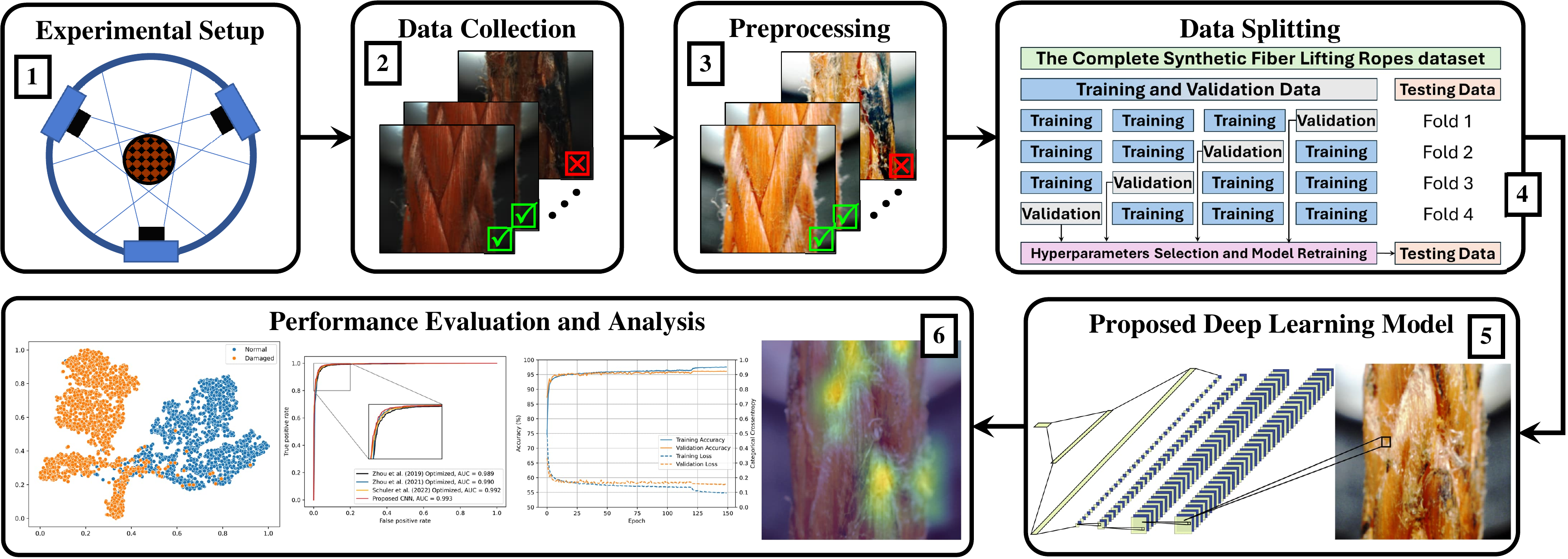}}
\caption{The proposed vision-based damage detection system for synthetic fiber lifting ropes.
The system is comprised of the following stages:
(1) experimental setup with a three-camera circular array to capture rope images;
(2) collection and annotation of the captured images;
(3) preprocessing to enhance quality and down-sampling to reduce complexity;
(4) data splitting into testing and training sets where the former is subdivided into 4-fold training and validation sets;
(5) training/testing the proposed deep learning model;
and (6) evaluating and analyzing the system's performance and computational complexity.}
\label{fig:system_model}
\end{figure*}
\begin{enumerate}
\item Setup an experimental apparatus with a three-camera circular array to photograph the ropes' surface area.
\item Collect the captured images and label them as normal or damaged according to the ropes' health condition.
\item Preprocess the collected images to enhance contrast and down-sample to reduce computational complexity.
\item Split the pre-processed images into train and test sets.
\item Use the train set to search for optimal parameters using 4-fold cross-validation and retrain with all train data.
\item Evaluate and analyze the model's performance using various metrics.
\end{enumerate}
The design process undertaken in this work is governed by the following requirements:
\begin{itemize}
\item High performance in detecting damaged ropes and robustness to different environmental and operational conditions.
\item Lightweight for implementation and deployment.
\item Remote sensing by neither interfering with the crane operation nor the rope structure.
\item Modularity to facilitate maintenance, upgrades, and compatibility with IoT and edge devices.
\end{itemize}
The remaining subsections discuss and detail each stage in the proposed system.
\subsection{Experimental setup}
The experimental setup was built and operated by Konecranes and the following experiment was repeated for three different synthetic fiber ropes; see Table \ref{table:properties} for the ropes' properties.
The crane illustrated in Fig. \ref{figure:crane} was set to continuously lift a payload of 5 metric tons in a controlled setting. The payload lifting height was approximately 5 meters and during a lifting cycle, the crane was stopped at the top and bottom (the payload resting on the floor). The process continued for weeks to cover the ropes' lifespan; from new to unusable.

\subsection{Imaging Apparatus}
We used a circular camera array comprising three RGB cameras placed at 120\textdegree\,apart, to capture approximately 13 mm of the lifting rope. The camera framerate was selected such that subsequent rope images have roughly 1/3 spatial overlap, resulting in 20 meters of rope being photographed during a lifting cycle. Further information, including an image of the apparatus, cannot be disclosed for patenting and trade secret reasons. However, schematics can be seen in the graphical abstract and Fig. \ref{fig:system_model}.
\begin{table}[!t]
\footnotesize
\centering
\caption{The fiber lifting rope properties.}
\begin{tabu}{cc}
\toprule
\textbf{Diameter} & 12 mm
\\\midrule
\textbf{Material} & Ultra high molecule weight polyethylene
\\\midrule
\textbf{Type} & 12-strand braided rope
\\\midrule
\textbf{Strength} & 15.4 metric tons (ISO 2307)
\\\midrule
\textbf{Weight} & 8.8 kg / 100 meters
\\\midrule
\textbf{Coating} & Abrasion and ultra-violet resistance
\\\bottomrule
\end{tabu}
\label{table:properties}
\end{table}
\subsection{Data collection}
\begin{figure}[!t]
\centerline{\includegraphics[width=0.9\columnwidth]{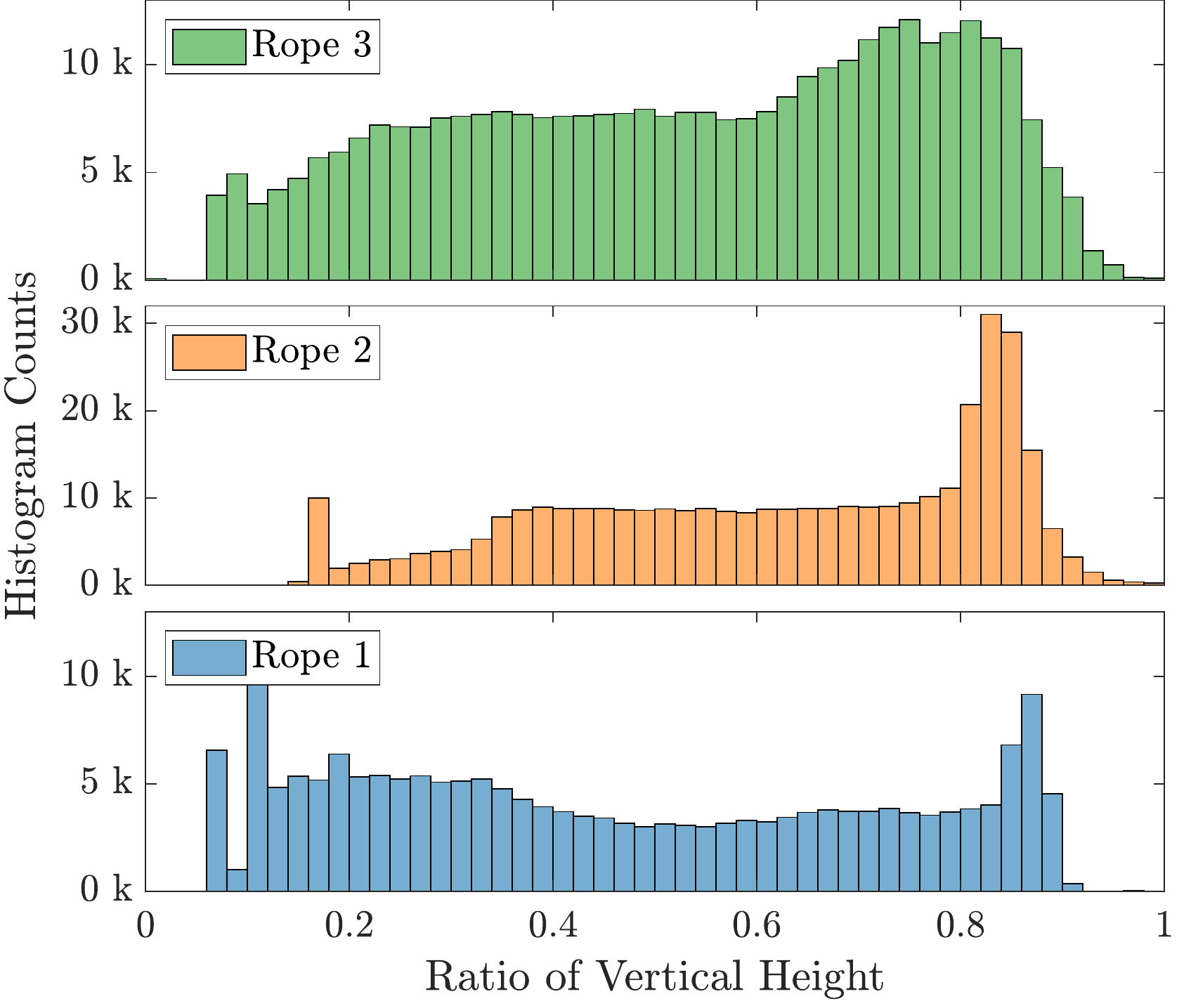}}
\caption{Histograms illustrating the distribution of rope height ratio for the 876,847 images from the three ropes.}
\label{fig:rope_histogram}
\end{figure}

The rope imaging experiments generated 4,984,000 high-resolution photos; each being tagged with a timestamp and the rope's imaged position.  The raw photos were then screened for duplicates by discarding images that examined the same rope position. In other words, we ensured that images for the same rope position would be distinct by capturing different health conditions. This is important to avoid cross-contamination in data splits. The 876,847 images from three ropes, categorized by rope height, are illustrated as histograms in Fig. \ref{fig:rope_histogram}.
After that, we selected 143,000 random samples and experts from Konecranes labeled them as \emph{normal} or \emph{damaged} according to the lifting rope health condition. Out of those images, 10,000 samples were labeled as damaged.
Finally, to avoid data imbalance issues, we formed a balanced dataset containing 20,000 samples; 10,000 images from each class.
The collected dataset is available from Konecranes and it was used under license for this study.
\subsection{Preprocessing and data splitting}

\begin{figure}[!t]
   \centering
   \subfloat[Raw sample. \label{fig:without_HE}]{\includegraphics[width=.475\columnwidth]{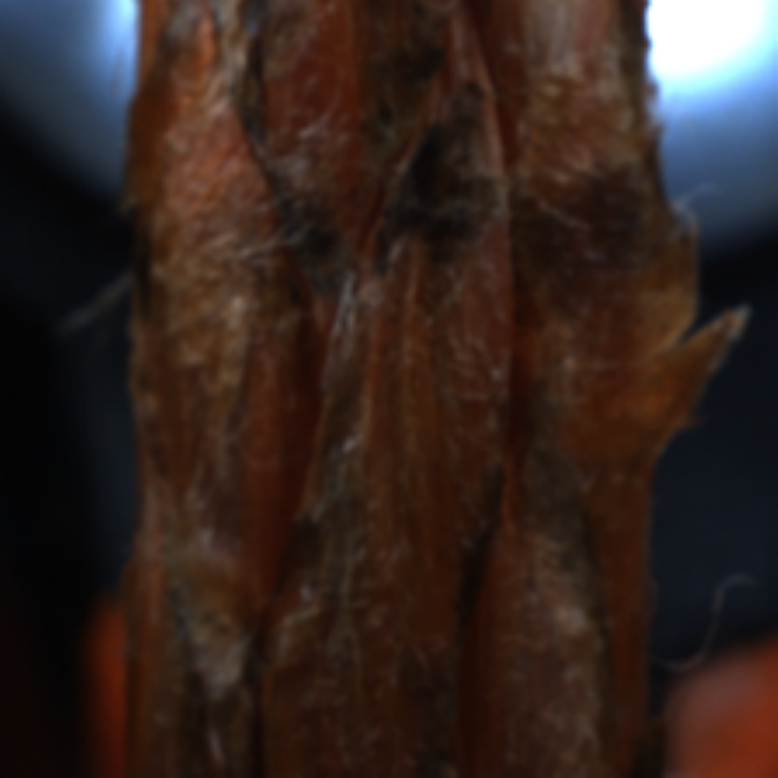}}
   \,\,\,\,
   \subfloat[Histogram equalized sample. \label{fig:with_HE}]{\includegraphics[width=.475\columnwidth]{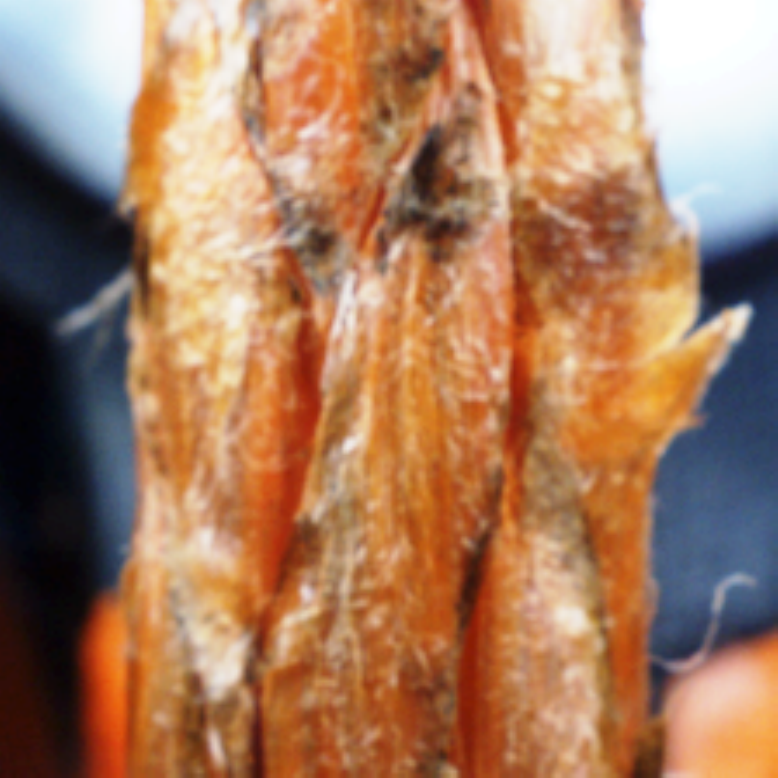}}
   \caption{Histogram equalization for an example rope image.}
   \label{fig:HE_example}
\end{figure}

The annotated high-resolution rope images were down-sampled to $256\times256\times3$ pixels. After that, we enhanced the photos' contrast via histogram equalization \cite{Mustafa_2018}; see Fig. \ref{fig:HE_example} for a sample, and we standardized the pixel values to range between 0 and 1. Finally, the pre-processed images were randomly divided into five equally sized portions while maintaining class balance; see stage 4 in Fig. \ref{fig:system_model}. One part (20 \%) was reserved as a testing set while the remaining four parts (80 \%) were utilized for training and validation using 4-fold stratified cross-validation method. In other words, training and validation had 16,000 (8,000 damaged and 8,000 normal) images and testing 4,000 (2,000 damaged and 2,000 normal) images. After the optimized parameters have been found using the cross-validation method, we combined the training and validation data to produce the final training set to be trained and evaluated against the test set.

\begin{figure*}[!t]
   \centering
   \subfloat[Damaged. \label{fig:cond_0}]{\includegraphics[width=.19\textwidth]{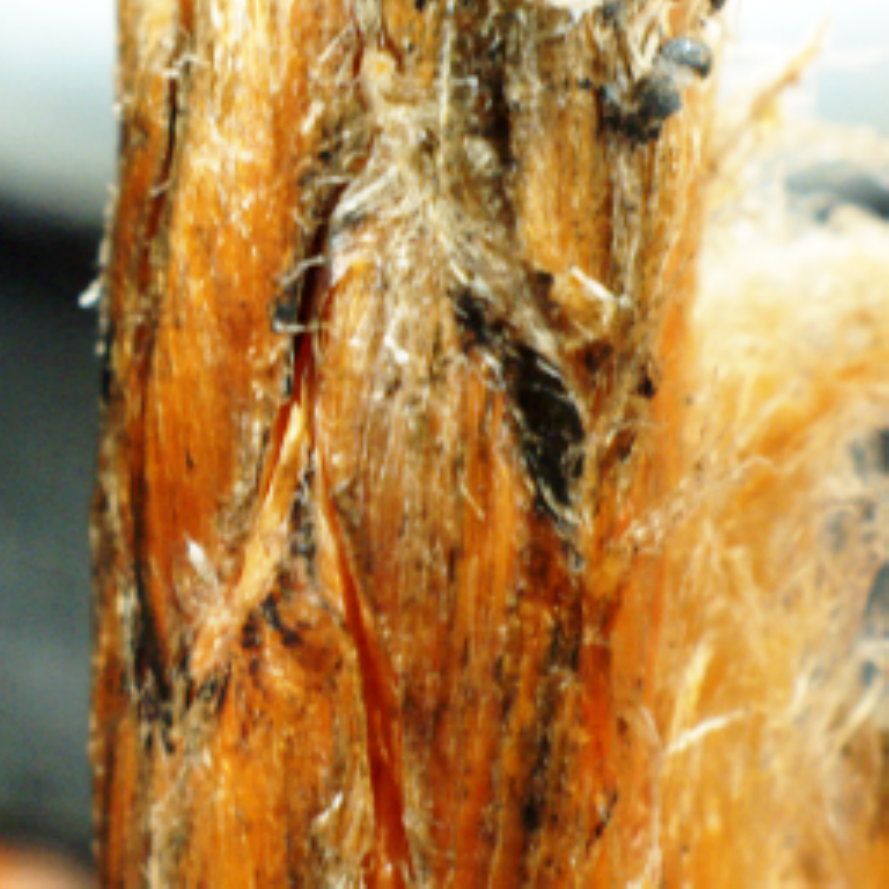}}
   \,
   \subfloat[Damaged. \label{fig:cond_1}]{\includegraphics[width=.19\textwidth]{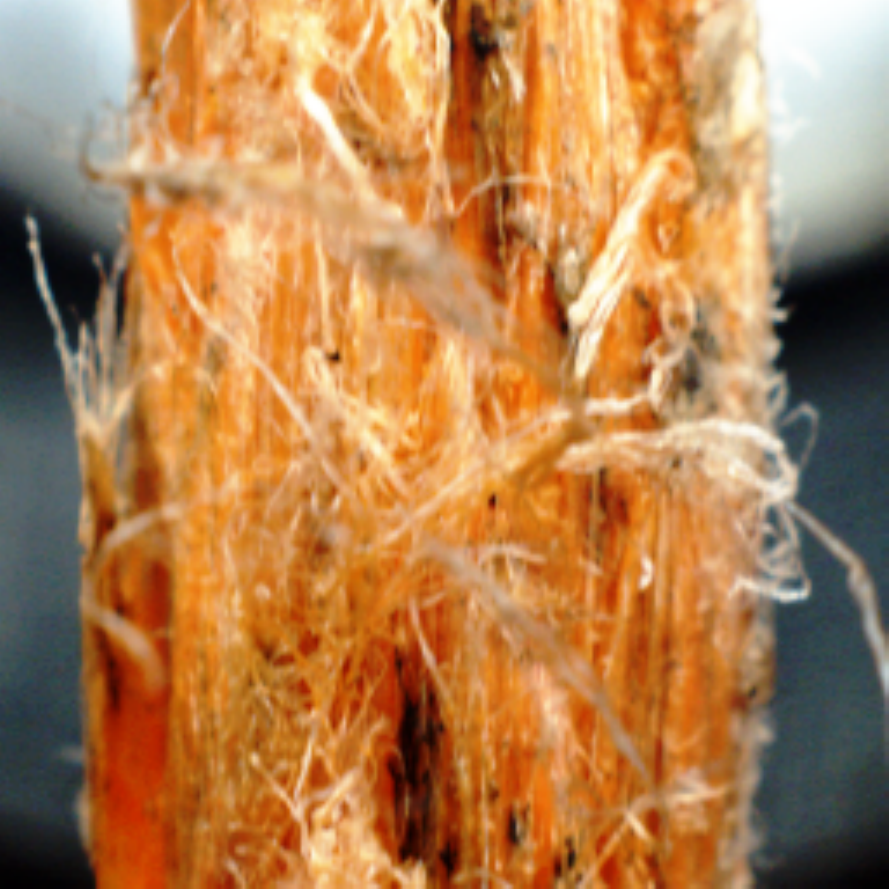}}
   \,
   \subfloat[Damaged. \label{fig:cond_2}]{\includegraphics[width=.19\textwidth]{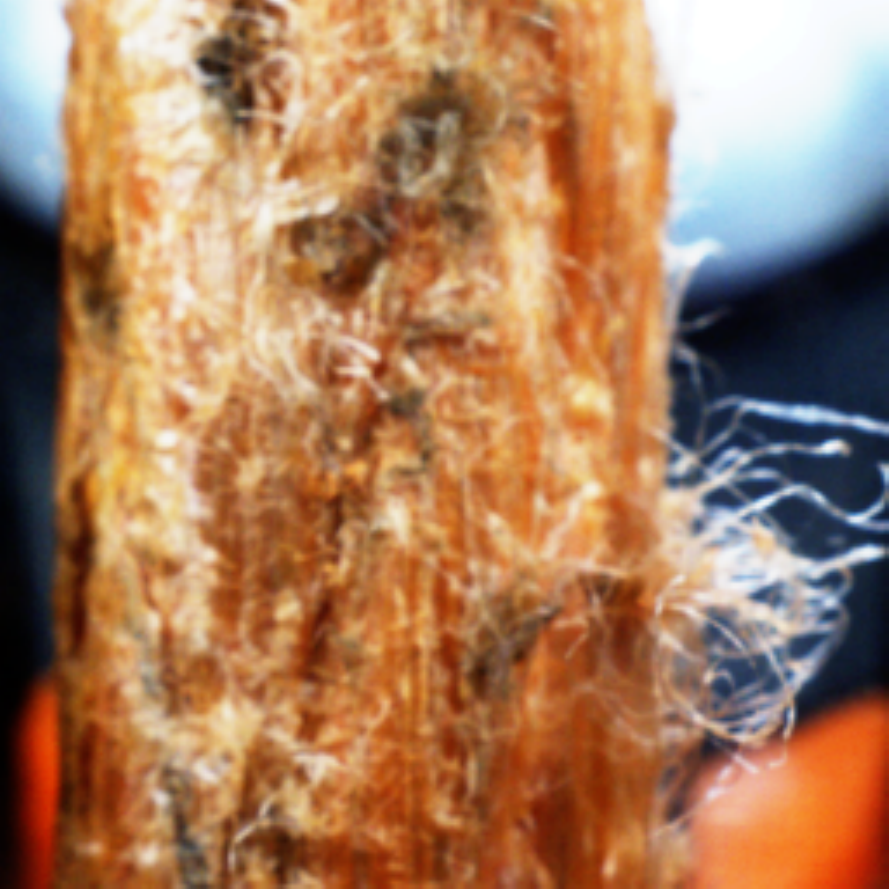}}
   \,
   \subfloat[Damaged. \label{fig:cond_3}]{\includegraphics[width=.19\textwidth]{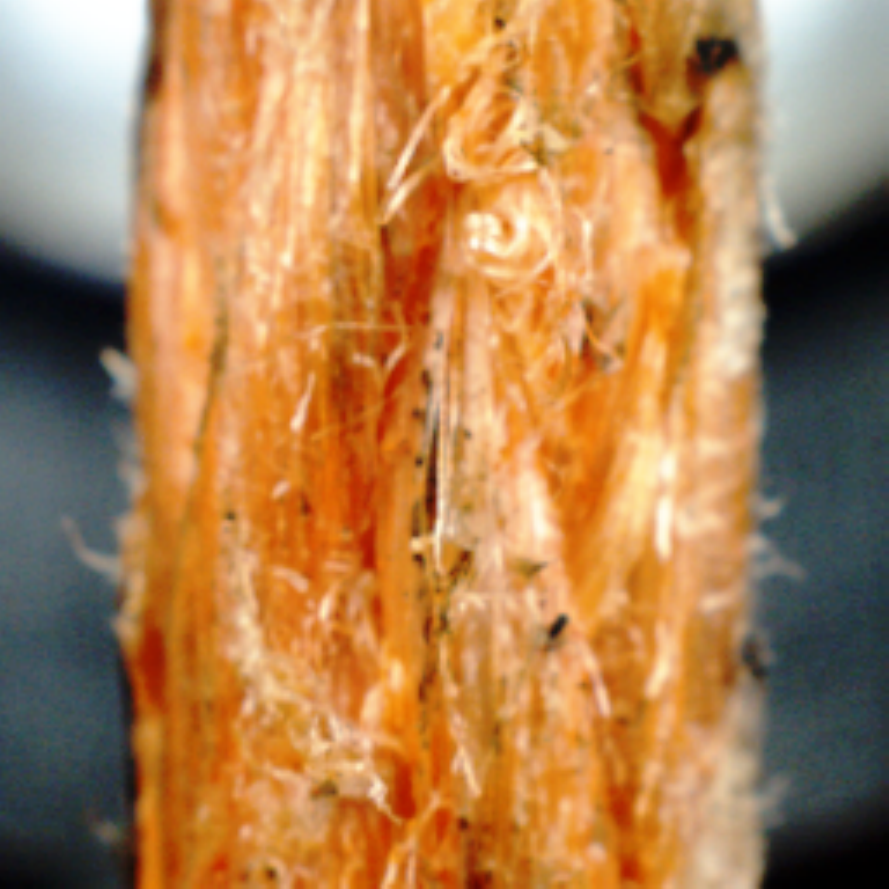}}
   \,
   \subfloat[Damaged. \label{fig:cond_4}]{\includegraphics[width=.19\textwidth]{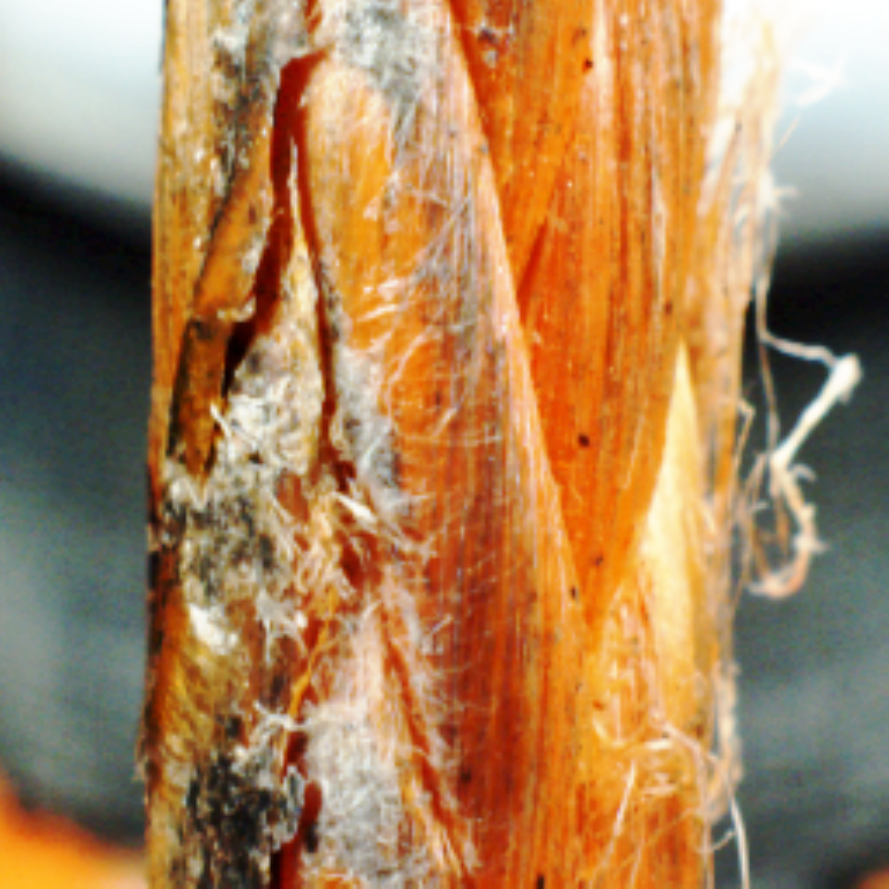}}
   \\
   \subfloat[Normal. \label{fig:cond_5}]{\includegraphics[width=.19\textwidth]{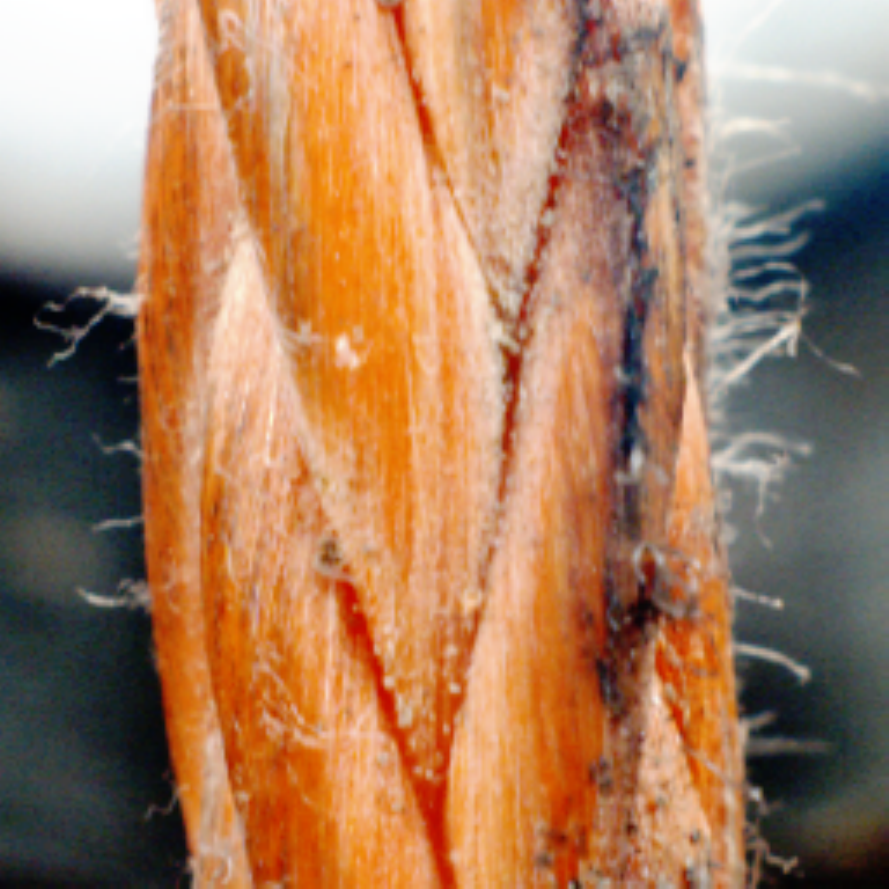}}
   \,
   \subfloat[Normal. \label{fig:cond_6}]{\includegraphics[width=.19\textwidth]{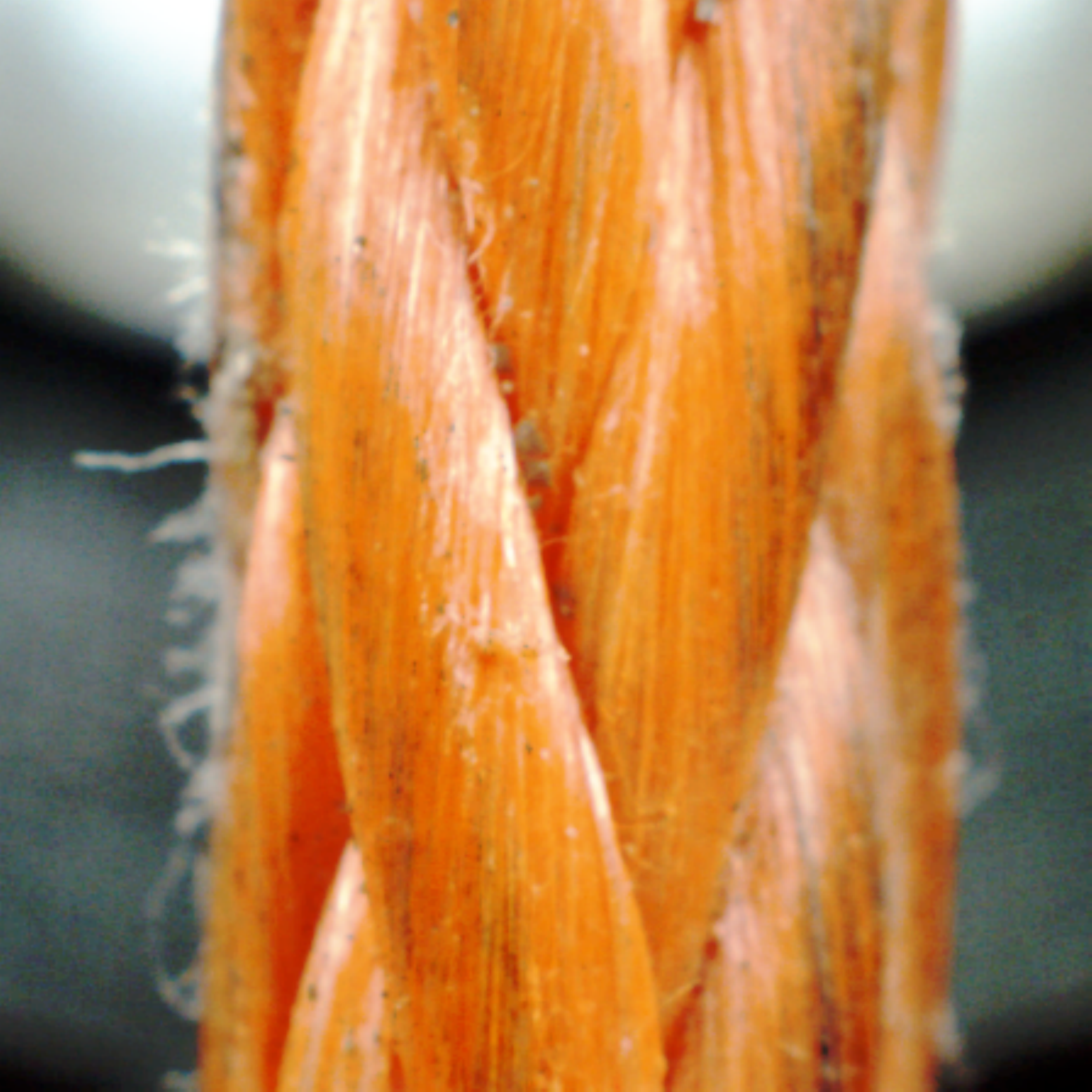}}
   \,
   \subfloat[Normal. \label{fig:cond_7}]{\includegraphics[width=.19\textwidth]{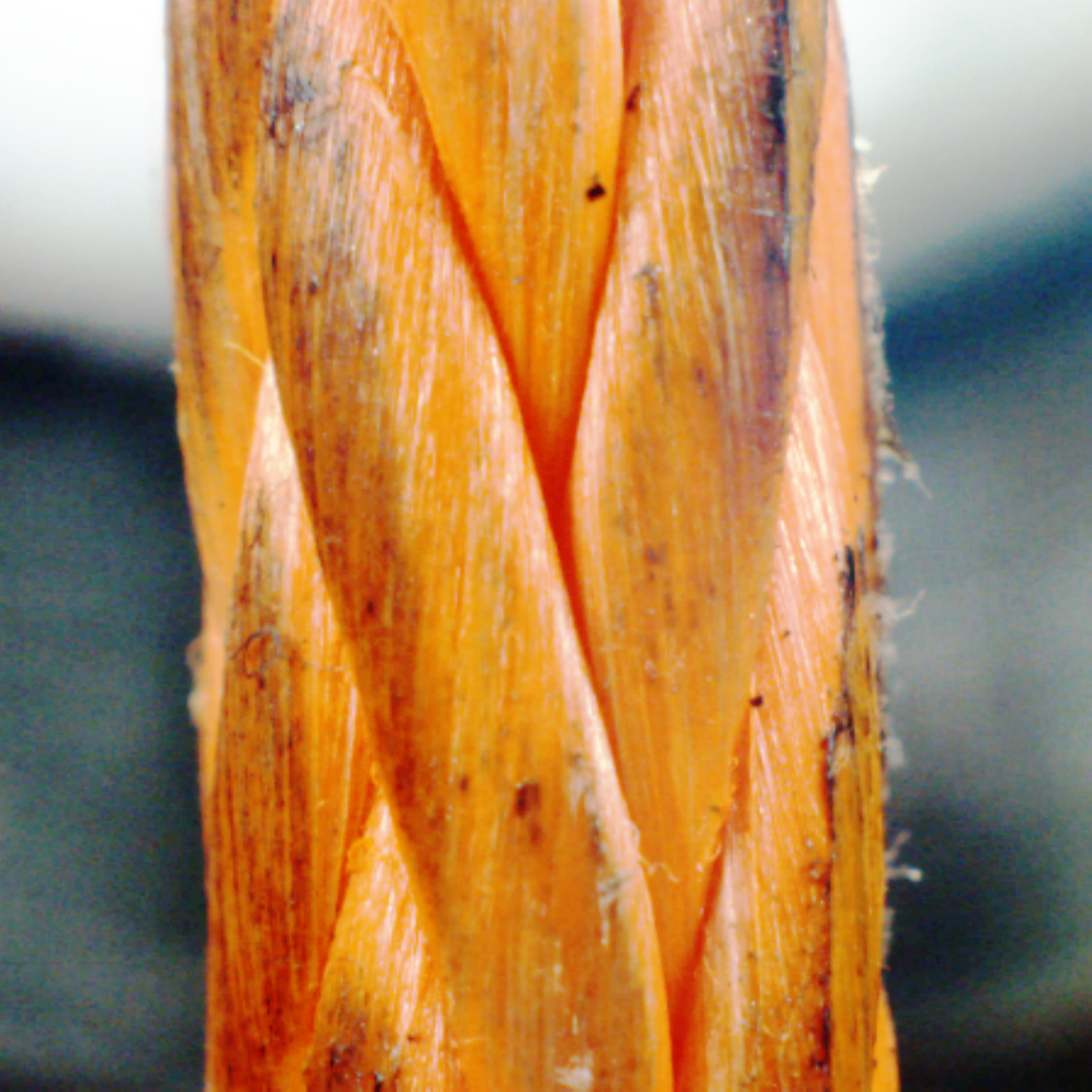}}
   \,
   \subfloat[Normal. \label{fig:cond_8}]{\includegraphics[width=.19\textwidth]{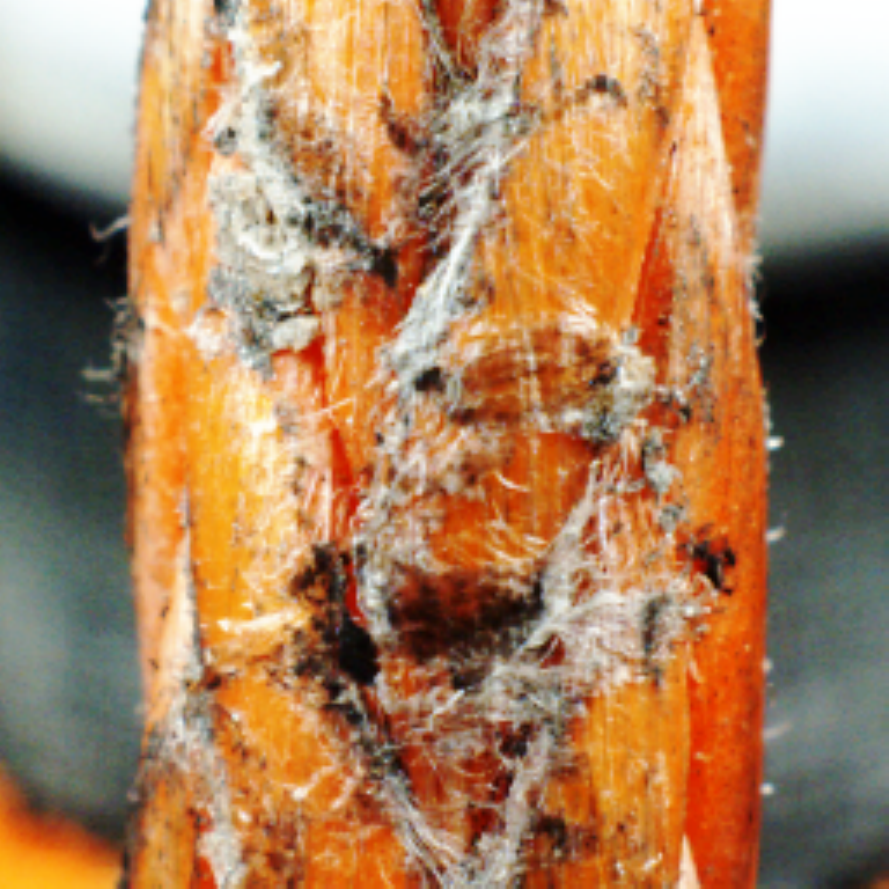}}
   \,
   \subfloat[Normal. \label{fig:cond_9}]{\includegraphics[width=.19\textwidth]{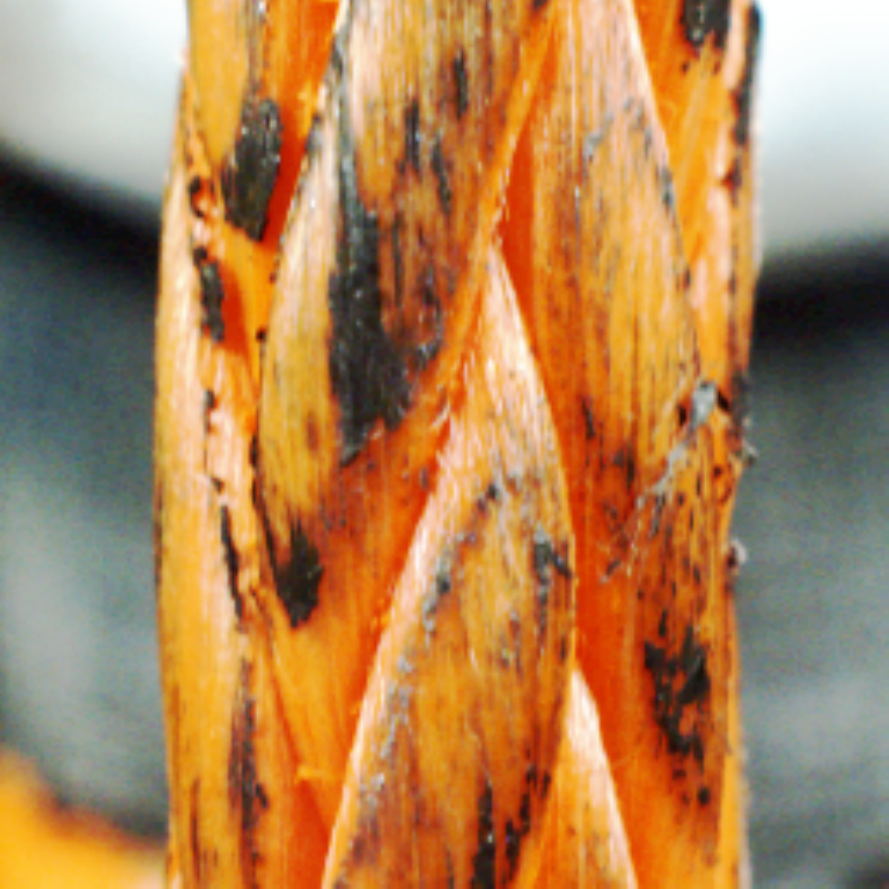}}
   \caption{Example images from the acquired dataset show significant variation in the severity and clarity of damages because of dirt and oil stains. The first row (a)-(e) shows damaged ropes while the second row (f)-(j) presents some healthy samples.}
   \label{fig:samples}
\end{figure*}

Fig. \ref{fig:samples} demonstrates samples of normal and damaged ropes from the acquired dataset. By examining the images, one notes a significant variation in the clarity of the rope's health state and in the severity of the damage. For example, Fig. \ref{fig:cond_0} conveys a more damaged rope when compared to the one presented in Fig. \ref{fig:cond_4}. However, the damage can also be minuscule without clear visual indications as presented in Fig. \ref{fig:cond_3}. Finally, the dirt and oil stains found in most rope images present a challenge for any vision-based tool.
\subsection{Proposed deep learning model} \label{sec:proposed_model}
The collected rope images constitute an over-complete description of the rope as a whole; hence, the problem of damage detection reduces to classifying each image independently. We designed a lightweight CNN architecture to classify the fiber lifting rope images, and we tested different variants to find the best-performing model.

The architecture design starts with a convolutional layer ($3\times3$ kernel with ReLU activation) to extract preliminary feature maps from the input images. After that, those initial features are passed through several blocks each consisting of the following sequential elements: (1) convolutional layer to extract features ($3\times3$ kernel with ReLU activation), (2) Max Pooling to down-sample the features ($2\times2$ kernel), (3) and dropout to regularize the network by reducing the neurons' interdependent learning ($0.4$ rate). Finally, the learned abstractions are flattened and passed through a dropout layer ($0.4$ rate), a fully connected layer ($20$ nodes), another dropout layer ($0.2$ rate), and lastly, a binary classification layer with a Softmax activation function. 
In this work, 16 model variants were generated from this architecture by altering the number of blocks (1, 2, or 3), input image sizes ($16\times16$, $32\times32$, or $64\times64$), and the input image color state (color or grayscale); see Appendix \ref{sec:model_variants} for the model variants' structure and Table \ref{table:ourcnn_structure} for details on the variant that we selected for further analysis and comparison.

The models' training was performed for 150 epochs using an Adam optimizer \cite{kingma2014adam} to minimize the cross-entropy loss regularized by a weight decay to reduce overfitting \cite{NEURIPS2018_f2925f97,van2017l2}, i.e.:
\begin{equation}
    \mathcal{L}= -y \log(\hat{y}) - (1-y) \log(1 - \hat{y}) + \lambda||\mathbf{w}||_2^2\,,
\end{equation}
where $\mathcal{L}$ denotes the regularized loss, $y$ and $\hat{y}$ are the true and predicted labels, respectively, $\lambda=5\times10^{-4}$ is the selected $L_2$ regularization rate, and $\mathbf{w}$ is the network's weight matrix \cite{NEURIPS2018_f2925f97}.
Moreover, the training batch size was set to 32 and to ensure convergence the learning rate was decayed by \cite{NEURIPS2019_2f4059ce}:
\begin{equation}
    \eta(n) = 
    \begin{cases}
    10^{-3} &: n \leq 120 \\
    10^{-4} &: 120 < n \leq 150 
    \end{cases}
    \,,
\end{equation}
where $\eta$ is the learning rate and $n$ is the epoch number.
\\
This training process was conducted for each model using the training set in each data fold (four training sets). Additionally, apart from the generated variants, we also trained the following three baseline models for comparison; Zhou \emph{et al.} (2019) \cite{Zhou2019}, Zhou \emph{et al.} (2021) \cite{Zhou2021}, and Schuler \emph{et al.} (2022) \cite{Schuler_2022}. We also optimized these models by training them with different image input sizes and parameters. Detailed descriptions of these models can be found in Appendices \ref{sec:baseline_model_1}-\ref{sec:baseline_model_2}.
\begin{table}[!t]
\footnotesize
\centering
\caption{The CNN9 model variant architecture.}
\begin{adjustbox}{max width=\columnwidth}
\begin{tabu}{ccccc}
\toprule
\textbf{Block \#} & \textbf{Layer} & \textbf{Information} & \textbf{Output shape} & \textbf{Parameters}
\\\midrule
\centering{\multirow{3}{*}{-}} & \multirow{3}{*}{Conv} & $L_2$ Kernel reg. $=0.0005$ & \multirow{3}{*}{$30\times30\times64$} & \multirow{3}{*}{1,792} \\
& & Kernel $=3\times3$ & & \\
& & Activation = ReLU & &
\\\midrule
\centering{\multirow{6.5}{*}{1}} & \multirow{3}{*}{Conv} & $L_2$ Kernel reg. $=0.0005$ & \multirow{3}{*}{$28\times28\times64$} & \multirow{3}{*}{36,928} \\
& & Kernel $=3\times3$ & & \\
& & Activation = ReLU & & 
\\\cmidrule{2-5}
& MaxPool & Pool $=2\times2$ & $14\times14\times64$ & 0 
\\\cmidrule{2-5}
& Dropout & Rate = 0.4 & $14\times14\times64$ & 0
\\\midrule
\centering{\multirow{6.5}{*}{2}} & \multirow{3}{*}{Conv} & $L_2$ Kernel reg. $=0.0005$ & \multirow{3}{*}{$12\times12\times64$} & \multirow{3}{*}{36,928} \\
& & Kernel $=3\times3$ & & \\
& & Activation = ReLU & &
\\\cmidrule{2-5}
& MaxPool & Pool $=2\times2$ & $6\times6\times64$ & 0
\\\cmidrule{2-5}
& Dropout & Rate = 0.4 & $6\times6\times64$ & 0
\\\midrule
\centering{\multirow{7}{*}{-}} & Flatten & - & 2,304 & 0
\\\cmidrule{2-5}
& Dropout & Rate = 0.4 & 2,304 & 0
\\\cmidrule{2-5}
& Dense & Activation = ReLU & 20 & 46,100
\\\cmidrule{2-5}
& Dropout & Rate = 0.2 & 20 & 0
\\\cmidrule{2-5}
& Dense & Activation = Softmax & 2 & 42
\\\midrule
\multicolumn{4}{c}{\textbf{Total number of parameters}} & \textbf{121,790}
\\\bottomrule
\end{tabu}
\end{adjustbox}
\label{table:ourcnn_structure}
\end{table}
\subsection{Performance evaluation and analysis} \label{sec:performance_evaluation}
The trained models were evaluated using the test set. Their performance was analyzed by various tools and metrics to quantify their detection, prediction, and misclassification outcomes.
\subsubsection{Classification}
we quantified the models' classification performance by accuracy, precision, recall, false positive rate (FPR), and the F1-score, i.e.:
\begin{equation}
\text{Accuracy} = \dfrac{TP+TN}{TP+TN+FP+FN}\,,
\end{equation}
\begin{equation}
\text{Precision} = \dfrac{TP}{TP+FP}\,,
\end{equation}
\begin{equation} \label{eq:recall}
\text{Recall} = \dfrac{TP}{TP+FN}\,,
\end{equation}
\begin{equation} \label{eq:fpr}
\text{FPR} = \dfrac{FP}{FP+TN}\,,
\end{equation}
\begin{equation}
\text{F1-score} = 2 \left(\dfrac{\text{Precision}\times\text{Recall}}{\text{Precision}+\text{Recall}}\right)\,,
\end{equation}
where $TP$, $TN$, $FP$, and $FN$ are true positives, true negatives, false positives, and false negatives, respectively (positive/negative denotes a damaged/normal rope).
\\
Moreover, we used the area under the receiver operating curve (AUC) and confusion matrices to fully characterize the classification quality. The AUC was computed using linearly interpolated receiver operating curves.
\subsubsection{Prediction}
we assessed the models' predictive capacity using Gradient-weighted Class Activation Mapping (Grad-CAM) which uses gradients of the last convolutional layer to measure the relevance of the input image pixels for classification \cite{selvaraju2017grad}. Specifically, Grad-CAM yields a distribution with high values for pixels that contributed more to the outcome.
\\
Furthermore, we utilized t-Distributed Stochastic Neighbor Embedding (t-SNE); a dimensionality reduction method that clusters similar high dimensional samples and departs dissimilar ones in two- or three-dimensional space \cite{van2008visualizing}. In specific, given an array of learned features $\mathbf{x}=[\mathbf{x}_1,\mathbf{x}_2,\cdots,\mathbf{x}_N]$, the similarity between features $i$ and $j$ can be measured by:
\begin{equation}
    p_{ij} = \frac{p_{j|i}+p_{i|j}}{2N}\,,
\end{equation}
\begin{equation}
    p_{j|i} =
    \begin{cases}
    \frac{\exp(-||\mathbf{x}_i-\mathbf{x}_j||^2/2\sigma_i^2)}{\sum_{k\neq i}\exp(-||\mathbf{x}_i-\mathbf{x}_k||^2/2\sigma_i^2)} &: i\neq j\\
    \qquad\qquad\quad0 &: i = j
    \end{cases}
    \,,
\end{equation}
where $p_{ij}$ is a probabilistic measure for the similarity between $\mathbf{x}_i$ and $\mathbf{x}_j$, $\sum_{i,j} p_{ij} = 1$, $\sum_j p_{j|i} = 1$, and $\sigma_i$ is the adaptive Gaussian kernel bandwidth.
Now, t-SNE aims to learn the two- or three-dimensional map $\mathbf{y}=[\mathbf{y}_1,\mathbf{y}_2,\cdots,\mathbf{y}_N]$ with a probabilistic similarity $q_{ij}$ that resembles $p_{ij}$, i.e. \cite{van2008visualizing}:
\begin{equation}
    q_{ij} = \frac{(1+||\mathbf{y}_i-\mathbf{y}_j||^2)^{-1}}{\sum_{k \neq l}(1+||\mathbf{y}_k-\mathbf{y}_l||^2)^{-1}}\,.
\end{equation}
The similarity matching between $q_{ij}$ and $p_{ij}$ in t-SNE is maximized by minimizing the Kullback–Leibler divergence of $p_{ij}$ from $q_{ij}$ via gradient descent, i.e.:
\begin{equation}
    \min_{\mathbf{y}_i} \left(\sum _{i\neq j}p_{ij}\log \left({\frac {p_{ij}}{q_{ij}}}\right)\right)\,.
\end{equation}
\\
Both the Grad-CAM and t-SNE help in characterizing the models' predictive power when supplied with new data. In other words, given large enough training samples, if the Grad-CAM and t-SNE show genuine learning and clear separability, one may infer the model adequacy for unseen samples.
\subsubsection{Misclassification}
visualizing the model's misclassified samples is paramount for interpretability and for outlining performance caps. Moreover, it enables a better understanding of the model's weaknesses and for identifying human errors in annotation. For example, by assuming some error in the labeling process, the performance of a genuine model will be limited, or capped, by the labels' quality \cite{6685834}.
\subsection{Computational complexity}\label{computational_complexity}
The complexity of the models was assessed by their total number of trainable parameters, required input image size, the models' memory size requirement, processing time, and their processing rate (frame rate). We analyzed their computational complexity by Monte-Carlo simulations where we fed the models with 1,000 test samples, predicted their health state (normal or damaged), and repeated the process ten times for validation. Note that this process does not include the imaging, data loading, nor preprocessing stages. It only quantifies the models' inference complexity.
We used an Apple MacBook Pro with an ARM-based M1 Pro chip, 10-core CPU, integrated 16-core GPU, 16-core neural engine, and 16 GB of RAM. The experiments' codes were written in Python 3 using Tensorflow 2 and are publicly available. 
\section{Results and discussion} \label{sec:results}
\subsection{Model selection}
The best model, out of the 16 generated variants, was selected based on its ability to balance between precision and recall with minimum computational requirements. By examining the results in Table \ref{table:perf_results_different_structures}, one notes that the model variants CNN9, CNN15, and CNN16 yield the highest 4-fold averaged precision/recall balance (96.3\% F1-score). Besides, they are the top 3 models in terms of accuracy, F1-score, and AUC. CNN16 has the best False Negative Rate (FNR) with 2.2\% while CNN9 is the second best with 2.7\%. Nevertheless, due to the apparent disparity in computational resources (input image sizes: $32\times32\times3$ v.s. $64\times64\times3$), we opted for model CNN9 and used it for further analysis and comparison.
Note that ``Proposed CNN" refers to the CNN9 variant in the remainder of this paper.
\begin{table}[!t]
\footnotesize
\centering

\caption{The proposed CNN model variants' validation performance in terms of accuracy, precision, recall, FNR, F1-score, and AUC. The results are summarized by their 4-fold averaged percentages, $\pm$ standard deviations, and the selected best model variant is highlighted in bold.}
\begin{adjustbox}{max width=\columnwidth}
\begin{tabu}{ccccccc}
\toprule
\textbf{Variant \#} & \textbf{Accuracy} & \textbf{Precision} & \textbf{Recall} & \textbf{FNR} & \textbf{F1-score} & \textbf{AUC}
\\\midrule
CNN1 & $94.5 \pm 0.4$ & $94.4 \pm 0.3$ & $94.6 \pm 0.9$ & $5.4 \pm 0.9$ & $94.5 \pm 0.4$ & $98.3 \pm 0.1$
\\\midrule
CNN2 & $93.5 \pm 0.2$ & $93.1 \pm 0.4$ & $93.9 \pm 0.4$ & $6.1 \pm 0.4$ & $93.5 \pm 0.2$ & $97.9 \pm 0.1$
\\\midrule
CNN3 & $95.8 \pm 0.2$ & $95.5 \pm 0.2$ & $96.1 \pm 0.4$ & $3.9 \pm 0.4$ & $95.8 \pm 0.2$ & $99.0 \pm 0.0$
\\\midrule
CNN4 & $95.5 \pm 0.3$ & $94.8 \pm 0.4$ & $96.1 \pm 0.2$ & $3.9 \pm 0.2$ & $95.5 \pm 0.3$ & $98.8 \pm 0.1$
\\\midrule
CNN5 & $95.5 \pm 0.1$ & $94.8 \pm 0.3$ & $96.3 \pm 0.3$ & $3.8 \pm 0.3$ & $95.5 \pm 0.1$ & $98.5 \pm 0.0$
\\\midrule
CNN6 & $95.5 \pm 0.1$ & $94.5 \pm 0.2$ & $96.7 \pm 0.1$ & $3.3 \pm 0.1$ & $95.6 \pm 0.1$ & $98.8 \pm 0.1$
\\\midrule
CNN7 & $93.2 \pm 2.3$ & $91.0 \pm 3.7$ & $96.1 \pm 0.1$ & $3.9 \pm 0.1$ & $93.4 \pm 2.0$ & $97.9 \pm 0.8$
\\\midrule
CNN8 & $96.0 \pm 0.2$ & $95.6 \pm 0.2$ & $96.4 \pm 0.4$ & $3.6 \pm 0.4$ & $96.0 \pm 0.2$ & $99.0 \pm 0.1$
\\\midrule
\textbf{CNN9} & $\mathbf{96.3 \pm 0.1}$ & $\mathbf{95.3 \pm 0.2}$ & $\mathbf{97.4 \pm 0.3}$ & $\mathbf{2.6 \pm 0.3}$ & $\mathbf{96.3 \pm 0.1}$ & $\mathbf{99.2 \pm 0.1}$
\\\midrule
CNN10 & $95.9 \pm 0.1$ & $94.8 \pm 0.3$ & $97.2 \pm 0.5$ & $2.8 \pm 0.5$ & $96.0 \pm 0.1$ & $99.0 \pm 0.1$
\\\midrule
CNN11 & $94.9 \pm 0.2$ & $94.7 \pm 0.5$ & $95.2 \pm 0.3$ & $4.8 \pm 0.3$ & $94.9 \pm 0.2$ & $98.2 \pm 0.1$
\\\midrule
CNN12 & $95.4 \pm 0.2$ & $94.6 \pm 0.6$ & $96.4 \pm 0.9$ & $3.6 \pm 0.9$ & $95.5 \pm 0.3$ & $98.8 \pm 0.2$
\\\midrule
CNN13 & $95.2 \pm 0.2$ & $93.5 \pm 0.4$ & $97.2 \pm 0.4$ & $2.8 \pm 0.4$ & $95.3 \pm 0.2$ & $98.8 \pm 0.1$
\\\midrule
CNN14 & $96.0 \pm 0.1$ & $95.6 \pm 0.4$ & $96.5 \pm 0.3$ & $3.5 \pm 0.3$ & $96.1 \pm 0.1$ & $98.9 \pm 0.1$
\\\midrule
CNN15 & $96.3 \pm 0.1$ & $95.3 \pm 0.3$ & $97.3 \pm 0.3$ & $2.7 \pm 0.3$ & $96.3 \pm 0.1$ & $99.1 \pm 0.0$
\\\midrule
CNN16 & $96.3 \pm 0.1$ & $94.9 \pm 0.3$ & $97.8 \pm 0.4$ & $2.2 \pm 0.4$ & $96.3 \pm 0.1$ & $99.2 \pm 0.0$
\\\bottomrule
\end{tabu}
\end{adjustbox}
\label{table:perf_results_different_structures}
\end{table}
\subsection{Performance analysis}

\begin{figure*}[!t]
   \centering
   \subfloat[Zhou \emph{et al.} (2019) Optimized. \label{fig:res_1}]{\includegraphics[width=.24\textwidth]{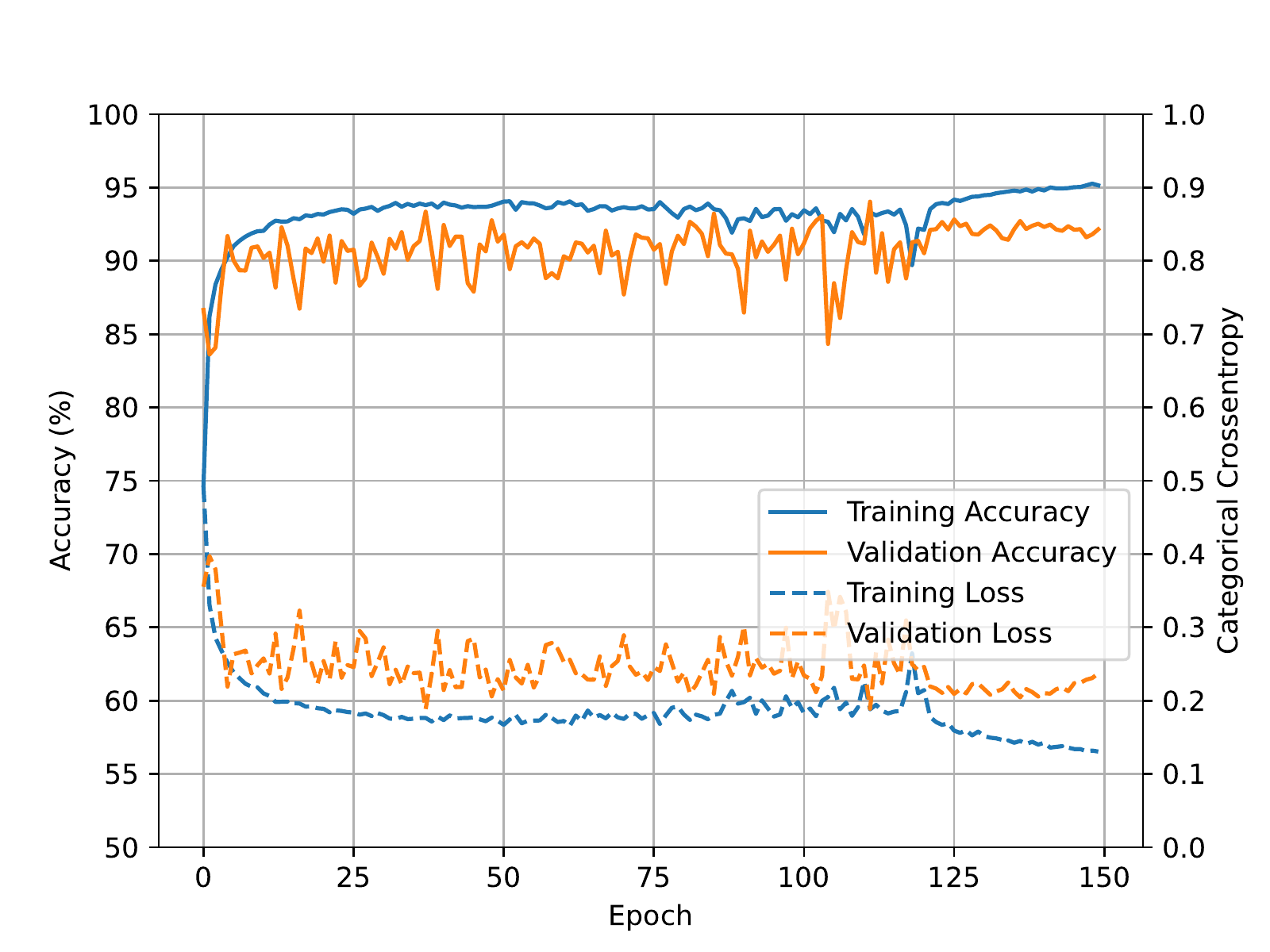}}\,
   \subfloat[Zhou \emph{et al.} (2021) Optimized. \label{fig:res_2}]{\includegraphics[width=.24\textwidth]{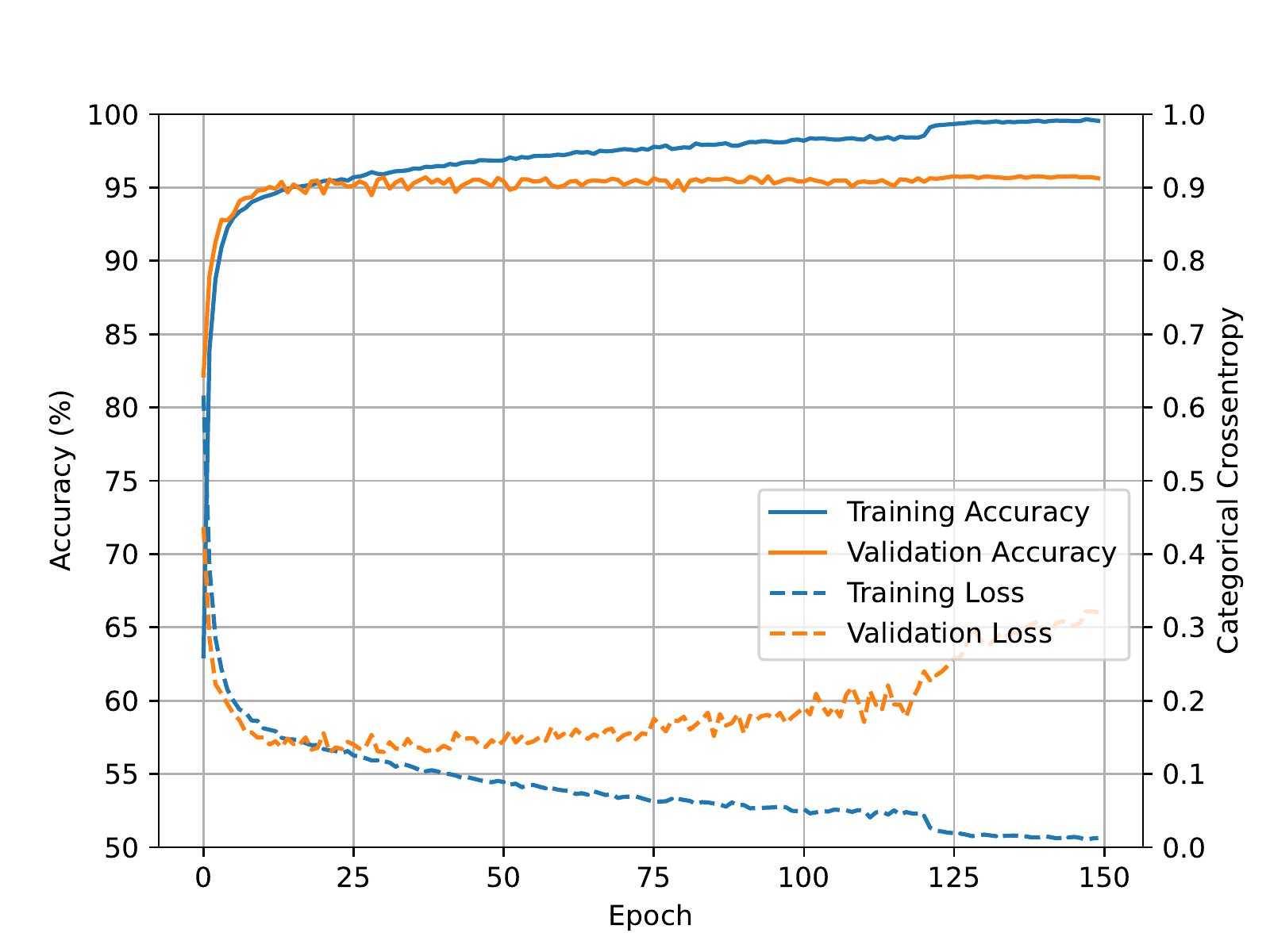}}\,
   \subfloat[Schuler \emph{et al.} (2022) Optimized. \label{fig:res_3}]{\includegraphics[width=.24\textwidth]{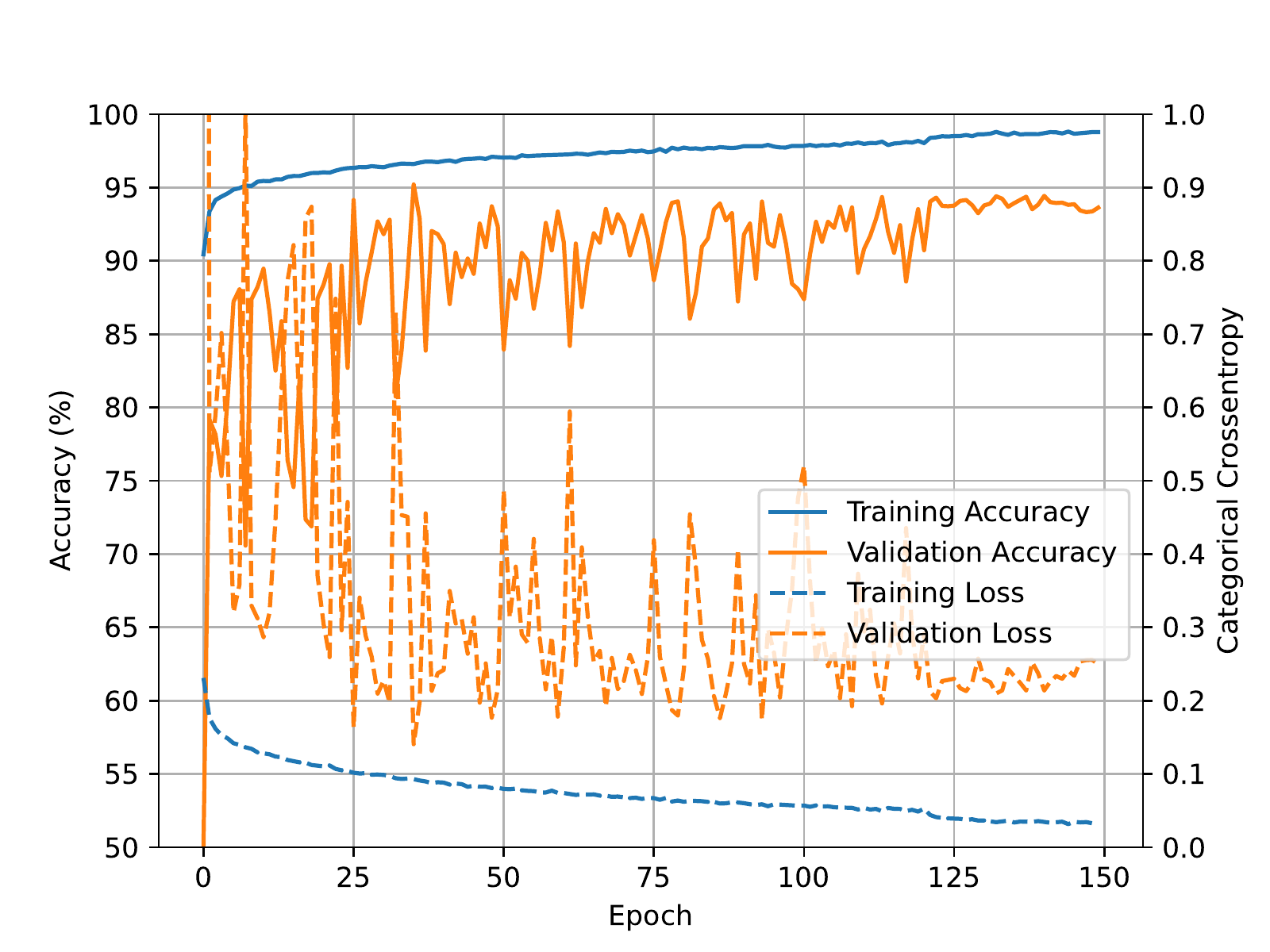}}\,
   \subfloat[Proposed CNN. \label{fig:res_4}]{\includegraphics[width=.24\textwidth]{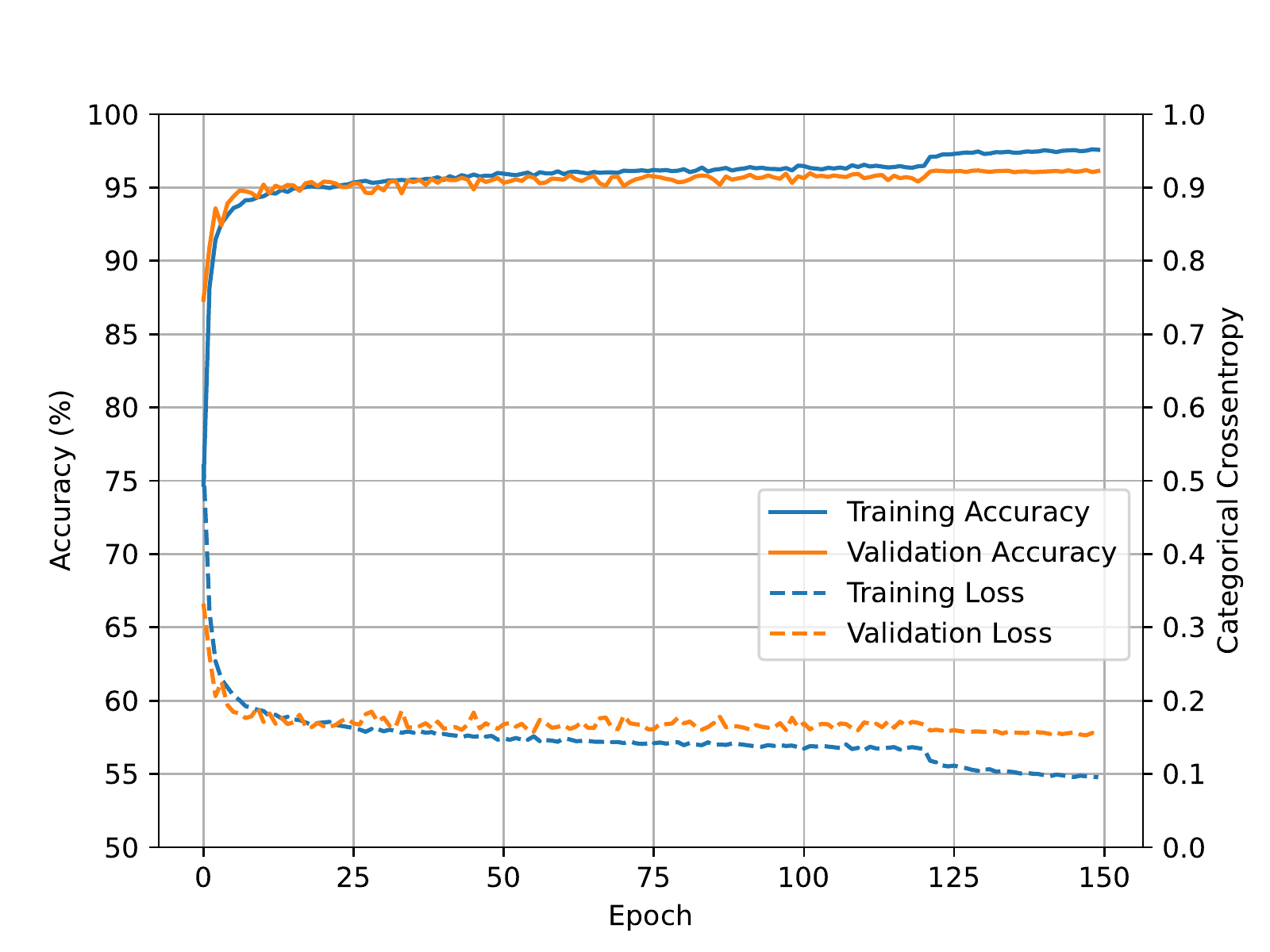}}
   \caption{Comparing the models' training and validation accuracy/loss curves averaged over the data folds in solid/dotted lines.}
   \label{fig:learning_acc}
\end{figure*}

Figure \ref{fig:learning_acc} compares the training and testing accuracy and loss curves (averaged over the data splits) of the proposed CNN with those of the three optimized baseline models.
The results show the curves converging successfully after epoch 120, and the learning rate decay scheduled there has ensured stability by suppressing perturbation. This indicates that the training phase was executed long enough and was not terminated prematurely. Moreover, comparing the differences between the training and validation accuracy and loss curves reveals that the optimized baseline models of Zhou \emph{et al.} (2019), Zhou \emph{et al.} (2021) and Schuler \emph{et al.} (2022) show greater overfitting compared to the proposed network.

\begin{figure}[!t]
\centerline{\includegraphics[width=0.9\columnwidth]{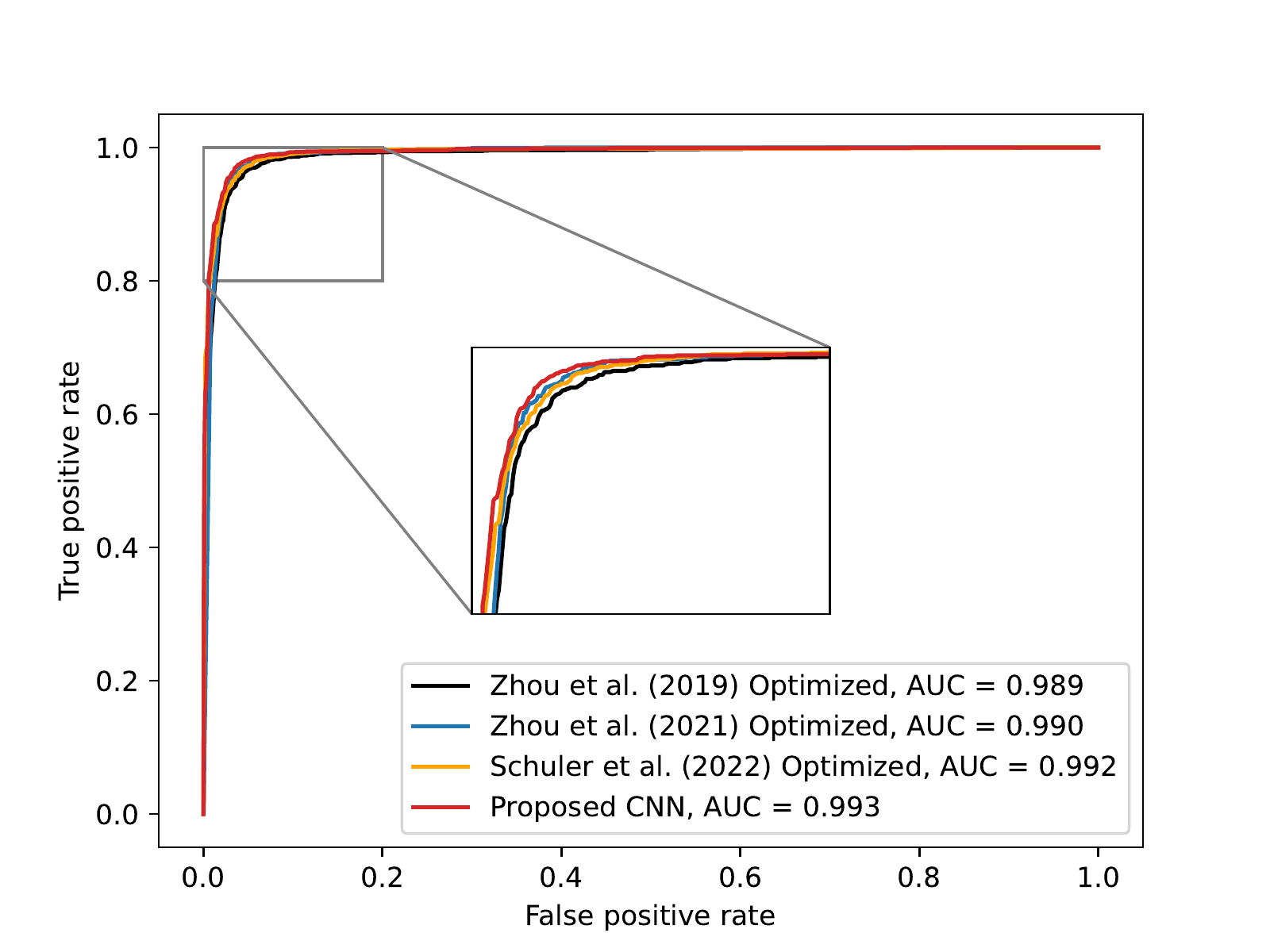}}
\caption{The models' ROC curves alongside their computed AUC values on testing data. The corner portion is magnified to ease visualization.}
\label{fig:roc}
\end{figure}

The original models' testing performance for detecting damaged rope images is demonstrated in Table \ref{table:tab_perf_results_original}, while Table \ref{table:tab_perf_results_optimized} and Fig. \ref{fig:roc} show the results after our parameter optimization. The measures in Table \ref{table:tab_perf_results_optimized} show that the optimized Schuler \emph{et al.} (2022) model has the highest recall at 99.7\%, which is the most important metric to prevent accidents. However, its precision, at 81.4\%, is extremely low suggesting that the model labels rope images almost systematically as damaged, making it unpractical for a real-life setting. Our Proposed CNN achieves the second best recall at 98.3\%. Moreover, both the optimized Zhou \emph{et al.} (2019) and Zhou \emph{et al.} (2021) models yield lower recall levels (97.2\% and 97.4\%, respectively). Nevertheless, the proposed CNN results in the highest accuracy and F1-score. In addition, it demonstrates the best precision/recall trade-off, which is reflected by the ROC curve in Fig. \ref{fig:roc} along with its AUC value (99.3\%). Also, our optimizations of Zhou \emph{et al.} (2019) and Zhou \emph{et al.} (2019) improved the performance of both models. Finally, the results suggest that the Zhou \emph{et al.} (2019) model is not the best in any of the six metrics.
\begin{table*}[!t]
\footnotesize
\centering
\caption{The original models' testing performance in terms of $TP$, $TN$, $FP$, and $FN$ presented in a confusion matrix fashion, along with their accuracy, precision, recall, FNR, and F1-score.
}
\begin{tabu}{p{1cm}p{1cm}cccc}
\toprule
& & \multicolumn{2}{c}{Zhou \emph{et al.} (2019) Original} & \multicolumn{2}{c}{Zhou \emph{et al.} (2021) Original}
\\\midrule
\multicolumn{1}{c|}{\centering{$\boldsymbol{TN}$}} & \centering{$\boldsymbol{FP}$} & \multicolumn{1}{p{1.5cm}|}{\centering{1,569}} & \multicolumn{1}{p{1.5cm}}{\centering{431}} & \multicolumn{1}{p{1.5cm}|}{\centering{1,875}} & \multicolumn{1}{p{1.5cm}}{\centering{125}}
\\\midrule
\multicolumn{1}{c|}{\centering{$\boldsymbol{FN}$}} & \centering{$\boldsymbol{TP}$} & \multicolumn{1}{p{1.5cm}|}{\centering{25}} & \multicolumn{1}{p{1.5cm}}{\centering{1,975}} & \multicolumn{1}{p{1.5cm}|}{\centering{67}} & \multicolumn{1}{p{1.5cm}}{\centering{1,933}}
\\\midrule
\multicolumn{2}{c}{\textbf{Accuracy}} 
& \multicolumn{2}{c}{88.6} 
& \multicolumn{2}{c}{95.2}
\\\midrule
\multicolumn{2}{c}{\textbf{Precision}} 
& \multicolumn{2}{c}{82.1}
& \multicolumn{2}{c}{93.9}
\\\midrule
\multicolumn{2}{c}{\textbf{Recall}} 
& \multicolumn{2}{c}{98.8}
& \multicolumn{2}{c}{96.7}
\\\midrule
\multicolumn{2}{c}{\textbf{FNR}} 
& \multicolumn{2}{c}{1.2}
& \multicolumn{2}{c}{3.4}
\\\midrule
\multicolumn{2}{c}{\textbf{F1-score}} 
& \multicolumn{2}{c}{89.7}
& \multicolumn{2}{c}{95.3}
\\\midrule
\multicolumn{2}{c}{\textbf{AUC}} 
& \multicolumn{2}{c}{98.2}
& \multicolumn{2}{c}{98.6}
\\\bottomrule
\end{tabu}
\label{table:tab_perf_results_original}
\end{table*}
\begin{table*}[!t]
\footnotesize
\centering
\caption{The optimized models' testing performance in terms of $TP$, $TN$, $FP$, and $FN$ presented in a confusion matrix fashion, along with their accuracy, precision, recall, FNR, and F1-score. 
}
\begin{tabu}{p{1cm}p{1cm}cccccccc}
\toprule
& & \multicolumn{2}{c}{Zhou \emph{et al.} (2019) Optimized} & \multicolumn{2}{c}{Zhou \emph{et al.} (2021) Optimized} & \multicolumn{2}{c}{Schuler \emph{et al.} (2022) Optimized} & \multicolumn{2}{c}{Proposed CNN}
\\\midrule
\multicolumn{1}{c|}{\centering{$\boldsymbol{TN}$}} & \centering{$\boldsymbol{FP}$} & \multicolumn{1}{p{1.5cm}|}{\centering{1,876}} & \multicolumn{1}{p{1.5cm}}{\centering{124}} & \multicolumn{1}{p{1.5cm}|}{\centering{\textbf{1,900}}} & \multicolumn{1}{p{1.5cm}}{\centering{\textbf{100}}} & \multicolumn{1}{p{1.5cm}|}{\centering{1,544}} & \multicolumn{1}{p{1.5cm}}{\centering{456}} & \multicolumn{1}{p{1.5cm}|}{\centering{1,893}} & \multicolumn{1}{p{1.5cm}}{\centering{107}}
\\\midrule
\multicolumn{1}{c|}{\centering{$\boldsymbol{FN}$}} & \centering{$\boldsymbol{TP}$} & \multicolumn{1}{p{1.5cm}|}{\centering{55}} & \multicolumn{1}{p{1.5cm}}{\centering{1,945}} & \multicolumn{1}{p{1.5cm}|}{\centering{52}} & \multicolumn{1}{p{1.5cm}}{\centering{1,948}} & \multicolumn{1}{p{1.5cm}|}{\centering{\textbf{6}}} & \multicolumn{1}{p{1.5cm}}{\centering{\textbf{1,994}}} & \multicolumn{1}{p{1.5cm}|}{\centering{34}} & \multicolumn{1}{p{1.5cm}}{\centering{1,966}}
\\\midrule
\multicolumn{2}{c}{\textbf{Accuracy}} 
& \multicolumn{2}{c}{95.5}
& \multicolumn{2}{c}{96.2}
& \multicolumn{2}{c}{88.5} 
& \multicolumn{2}{c}{\textbf{96.5}}
\\\midrule
\multicolumn{2}{c}{\textbf{Precision}} 
& \multicolumn{2}{c}{94.0}
& \multicolumn{2}{c}{\textbf{95.1}}
& \multicolumn{2}{c}{81.4} 
& \multicolumn{2}{c}{94.8}
\\\midrule
\multicolumn{2}{c}{\textbf{Recall}} 
& \multicolumn{2}{c}{97.2}
& \multicolumn{2}{c}{97.4} 
& \multicolumn{2}{c}{\textbf{99.7}} 
& \multicolumn{2}{c}{98.3}
\\\midrule
\multicolumn{2}{c}{\textbf{FNR}} 
& \multicolumn{2}{c}{2.8}
& \multicolumn{2}{c}{2.6} 
& \multicolumn{2}{c}{\textbf{0.3}} 
& \multicolumn{2}{c}{1.7}
\\\midrule
\multicolumn{2}{c}{\textbf{F1-score}} 
& \multicolumn{2}{c}{95.6} 
& \multicolumn{2}{c}{96.2} 
& \multicolumn{2}{c}{89.6} 
& \multicolumn{2}{c}{\textbf{96.5}}
\\\midrule
\multicolumn{2}{c}{\textbf{AUC}} 
& \multicolumn{2}{c}{98.9} 
& \multicolumn{2}{c}{99.0} 
& \multicolumn{2}{c}{99.2} 
& \multicolumn{2}{c}{\textbf{99.3}}
\\\bottomrule
\end{tabu}
\label{table:tab_perf_results_optimized}
\end{table*}
\subsection{Computational complexity analysis}
Table \ref{table:comp_results} summarizes the complexity analysis results which indicate that the proposed model is the fastest one with our optimized Zhou \emph{et al.} (2019) requiring approximately 30 milliseconds per input image and running in real-time at 33 fps. Interestingly, the optimized Zhou \emph{et al.} (2019) reaches the same processing speed with 5 million parameters as our Proposed CNN with 122 K parameters. This suggests the the equipment is not limited by the number of parameters although the effect in training time is apparent (27 seconds vs. 15 seconds per epoch). Nonetheless, it is important to note that the optimized Zhou \emph{et al.} (2021) model is comparatively fast, but the Schuler \emph{et al.} (2022) operates below real-time at 8 fps. Besides, the listed prediction speeds could be further improved by running them in C++. However, the most notable and important difference is the proposed model's light memory footprint. In specific, our model accepts small-sized images and requires less disk space for storage. These advantages can lead to savings in equipment and operational costs, improve latency.
\begin{table*}[!t]
\footnotesize
\centering
\caption{The models' computational complexity in terms of their total number of parameters, input image size requirement, model size, average processing time $\pm$ standard deviation in milliseconds, average processing rate in fps, and average training time per epoch $\pm$ standard deviation in seconds.}
\begin{tabu}{ccccc}
\toprule
\multicolumn{1}{c}{} & Zhou \emph{et al.} (2019) Optimized & Zhou \emph{et al.} (2021) Optimized & Schuler \emph{et al.} (2022) Optimized & Proposed CNN
\\\midrule
\textbf{No. of parameters} & 5.0 M & 268 K & \textbf{83 K} & 122 K
\\\midrule
\textbf{Image size} & \textbf{2.8 KB} & 10.3 KB & \textbf{2.8 KB} & \textbf{2.8 KB}
\\\midrule
\textbf{Model size} & 60.1 MB & \textbf{1.3 MB} & 4.6 MB & 1.7 MB
\\\midrule
\textbf{Processing time} & \textbf{30.3 $\pm$ 3.6 ms} & 30.7 $\pm$ 3.6 ms & 120.5 $\pm$ 9.1 ms & \textbf{30.3 $\pm$ 3.4 ms}
\\\midrule
\textbf{Processing rate} & \textbf{33.0 fps} & 32.6 fps & 8.3 fps & \textbf{33.0 fps}
\\\midrule
\textbf{Training time per epoch} & 26.5 $\pm$ 2.4 s & 25.2 $\pm$ 0.9 s & 217.9 $\pm$ 5.4 s & \textbf{15.2 $\pm$ 0.2 s}
\\\bottomrule
\end{tabu}
\label{table:comp_results}
\end{table*}
\\
\subsection{The Grad-CAM and t-SNE analysis}
\begin{figure}[!t]
   \centering
   \subfloat[Example 1. \label{fig:heatmap_0}]{\includegraphics[width=.475\columnwidth]{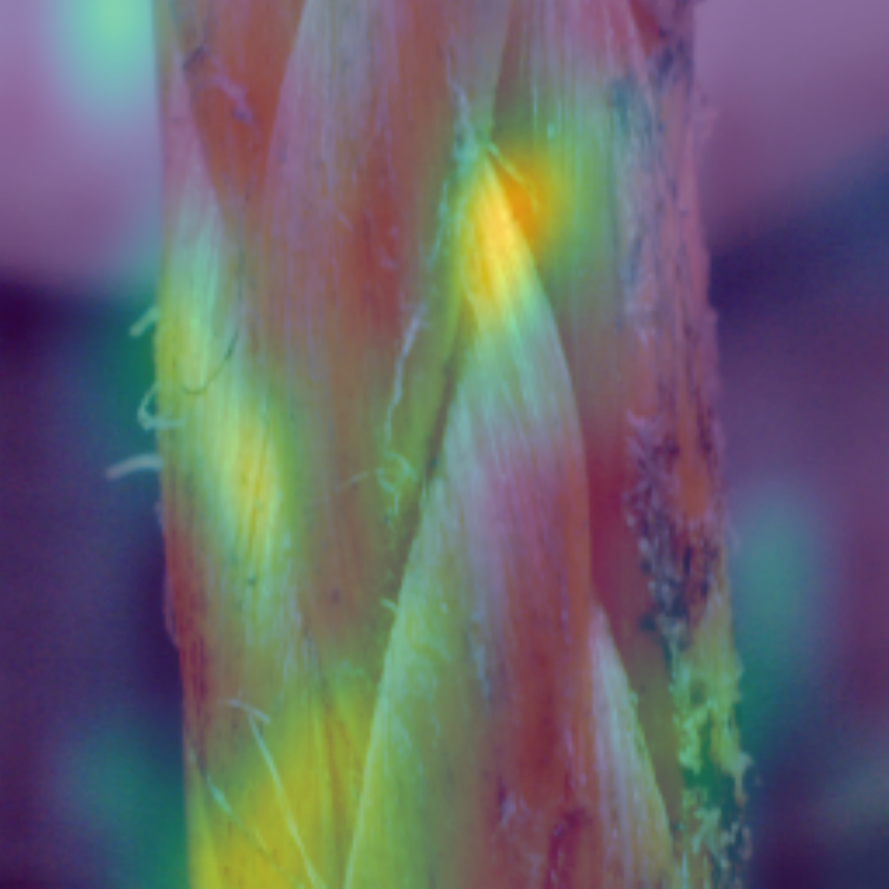}}
   \,\,\,\,
   \subfloat[Example 2. \label{fig:heatmap_1}]{\includegraphics[width=.475\columnwidth]{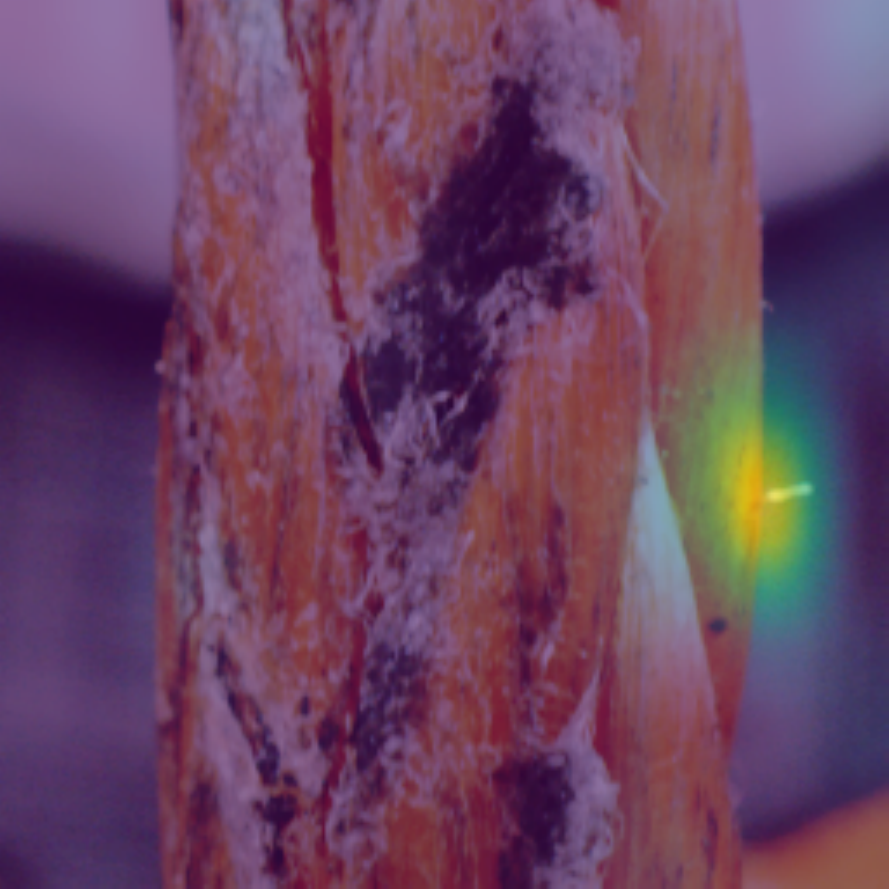}}
    \caption{The Proposed CNN Grad-CAM heatmaps for two correctly classified damaged ropes. The heatmaps indicate the model's adequacy by focusing on the pixels that are relevant for detecting damage. We used the model's last convolutional layer features as input to the Grad-Cam algorithm.}
    \label{fig:heatmap}
\end{figure}

\begin{figure}[!t]
\centerline{\includegraphics[width=0.9\columnwidth]{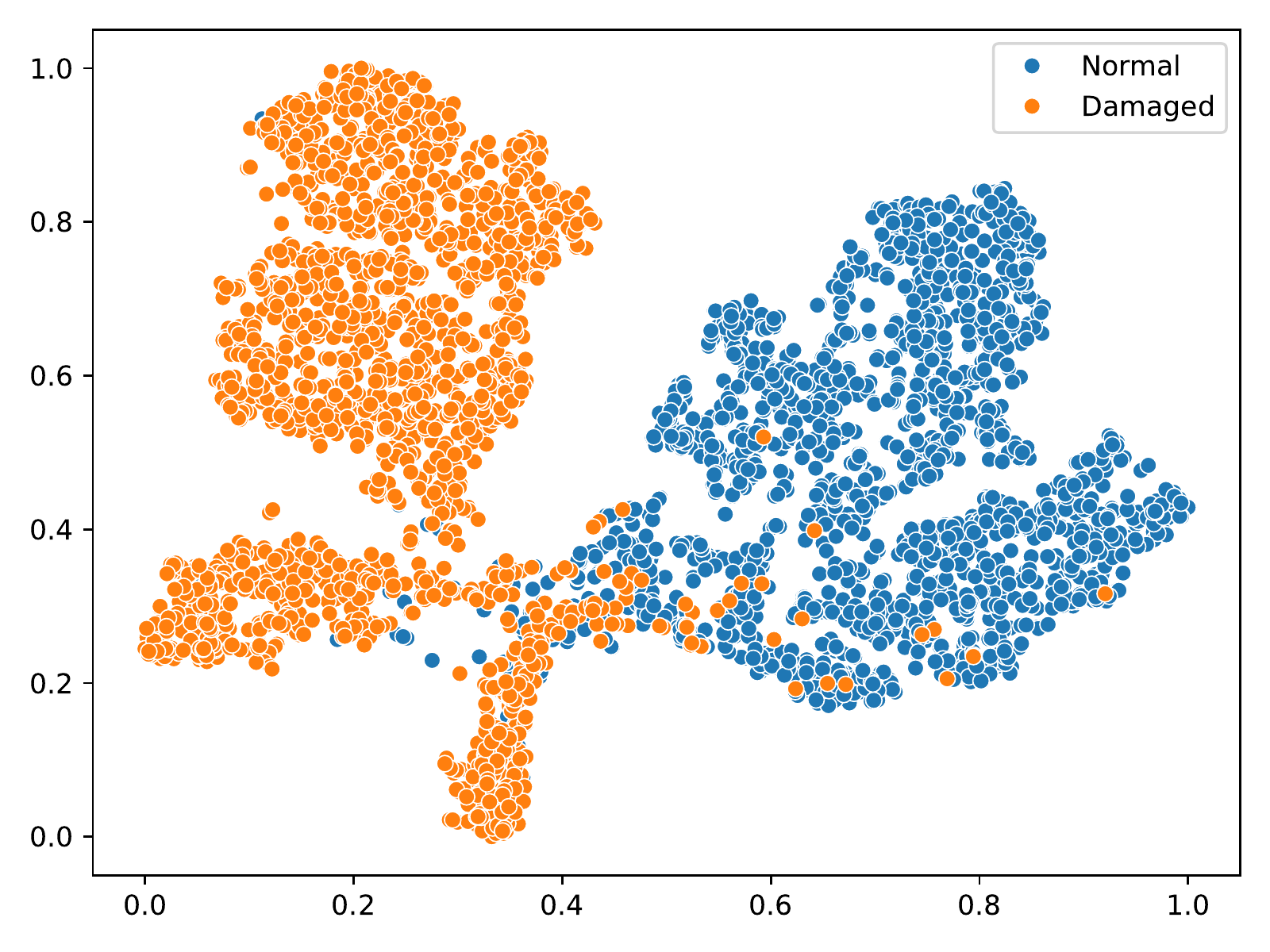}}
\caption{The Proposed CNN t-SNE testing set results show good separation between the two classes. We used the model's last layer features (FC[20] in Table \ref{table:proposed_cnn_alternatives}) as input to the t-SNE algorithm.}
\label{fig:t_sne}
\end{figure}

The Grad-CAM and t-SNE results are depicted in Figs. \ref{fig:heatmap} and \ref{fig:t_sne}, respectively.
In Fig. \ref{fig:heatmap}, we generated the Grad-CAM heatmap for two example images showing damaged ropes that were correctly classified by the proposed model. The results show that our CNN model is indeed focusing on the intuitively relevant parts of the input image, which are the broken strands. One also notes that the network does not focus on the ropes' oil and dirt residue as demonstrated in Fig. \ref{fig:heatmap_1}. This suggests genuine learning by the model and robustness to environmental and operational conditions.
However, it is important to note that the CNN still slightly focuses on the image background.
Moreover, the t-SNE results in Fig. \ref{fig:t_sne} demonstrate good class separation, but they also show a need for verifying some ground-truth labels. In specific, the t-SNE shows few rope images labeled as normal within the damaged rope support and vice versa.
\subsection{Misclassifications}
\begin{figure}[!t]
   \centering
   \subfloat[\scriptsize{True: Damaged, Predicted: Normal}. \label{fig:misclassified_0}]{\includegraphics[width=.475\columnwidth]{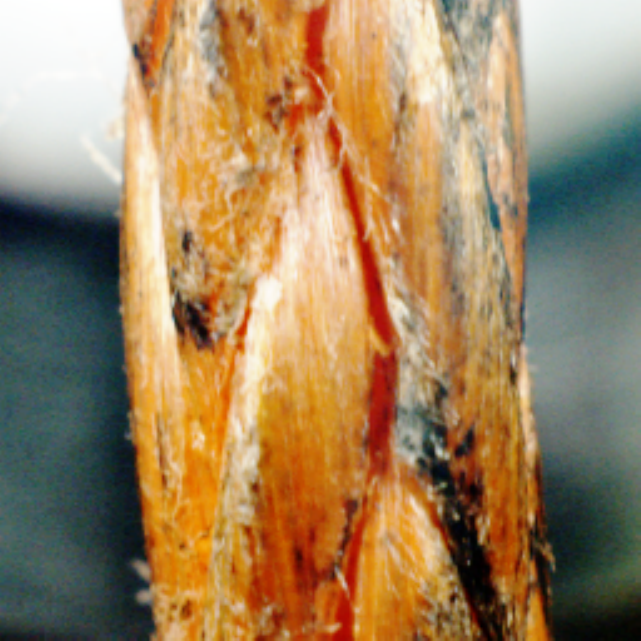}}
   \,\,\,\,
   \subfloat[\scriptsize{True: Normal, Predicted: Damaged}. \label{fig:misclassified_1}]{\includegraphics[width=.475\columnwidth]{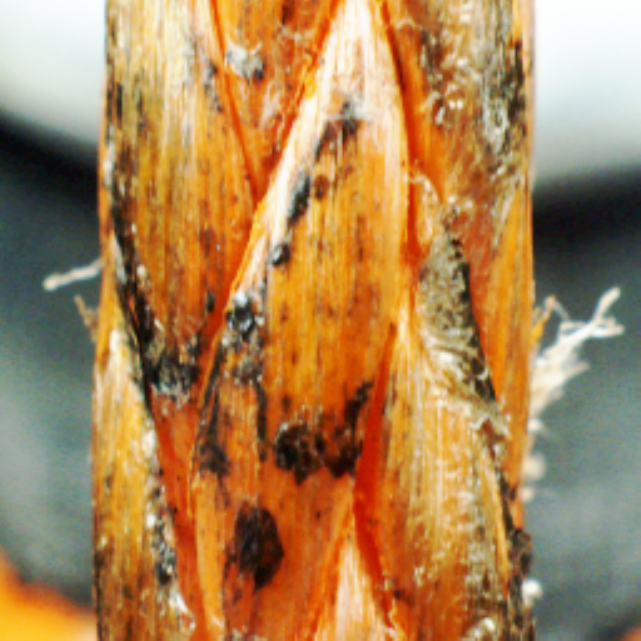}}
   \caption{Two example misclassification samples by the Proposed CNN.}
   \label{fig:misclassified}
\end{figure}

Two example prediction errors by our CNN are presented in Fig. \ref{fig:misclassified}. The rope in Fig. \ref{fig:misclassified_0} is labeled as damaged, but predicted as normal, while the one in Fig. \ref{fig:misclassified_1} is labeled as normal, but predicted as damaged. Such instances pose a challenge for the system because they are clearly in between the two classes; they are slightly worn with a few broken strings, but strictly not damaged, according to our experts. However, the rope in \ref{fig:misclassified_1} could be damaged on the other side of the rope, but it is hard to determine from this angle. Despite that, these ropes are not likely to break at these spots and they would be classified as damaged after more wear. Moreover, the similarities between the two images suggest possible annotation errors that may prevent the proposed model from reaching its full potential. Nonetheless, human errors are expected, and the model outcome still shows good potential and applicability.
\section{Conclusions} \label{sec:conclusions}
Damaged lifting ropes are a major safety hazard in transportation, manufacturing, cargo loading/unloading, and construction because they can lead to serious accidents, injuries, and financial losses.
Synthetic lifting ropes have many benefits over traditional steel wire ropes. Nonetheless, like steel wires, they are subject to wear and tear and can become damaged over time. Common types of damage in fiber ropes include strand cuts, abrasion, melting, compression damage, pulled strands, and variations in diameter. Fortunately, unlike steel wires, which often fail internally, damages in synthetic ropes typically manifest on the surface and can be detected through visual inspection. However, the visual inspection of damage in synthetic lifting ropes is a time-consuming task, interrupts operation, and may result in the premature disposal of ropes. Therefore, combining computer vision and deep learning techniques is intuitive for automation and advancement.

This work presents a vision-based deep learning solution for detecting damage in fiber lifting rope images.
First, we built a three-camera circular array apparatus to photograph the rope's surface. Afterward, the rope surface images were collected in a database, annotated by experts, preprocessed to improve contrast and split into training and validation sets with a separate testing set. The training and validation sets were used in a 4-fold cross-validation training setup to find optimal parameters.
Moreover, we systematically designed an efficient CNN-based model to classify damaged rope images, evaluated its detection and prediction performance using various tools, and compared it to three different baseline models. Additionally, we analyzed its computational complexity in terms of processing time and memory footprint.
In summary, the results indicated various performance and computational advantages for using the proposed system when compared to similar solutions.
Specifically, the system testing yielded 96.5\% accuracy, 94.8\% precision, 98.3\% recall, 96.5\% F1-score, 99.3\% AUC, and a significant generalization capability. Besides, it runs at 33 fps, occupies 1.7 MB of memory, and requires low-resolution input images; thus, making the proposed system a real-time lightweight solution.
The developed system was also found robust to various environmental and operational conditions e.g., oil residue and dust, and showed potential for implementation in diverse industrial applications that utilize fiber ropes, including but not limited to mooring, towing, sailing, and climbing.

The proposed model's main drawback is not determining the rope health state as a whole but assessing each image individually. In addition, since the model detects only surface anomalies, it may miss internal damages such as fiber degradation or internal strand breaks.
Besides, the system's output is binary and does not directly communicate the rope's health condition. Moreover, the proposed convolutional neural network is not the most modern machine learning technique. These limitations propose extending the developed solution in various ways such as:
(1) collecting a larger training dataset with different rope sizes, types, and payloads to improve generalization;
(2) investigating other more modern machine learning solutions and techniques;
(3) including the cost of the imaging apparatus in the design process, e.g., using cheaper cameras;
(4) extending the model's output to multiple classes e.g., normal, worn, and damaged, or to a continuous score indicating the rope's health condition (regression);
and (5) incorporating the proposed solution to automate or recommend spare-part ordering.
\section*{Acknowledgment}
We would like to thank Konecranes and Juhani Kerovuori for their collaboration on this project. This work was supported in part by Konecranes Plc through the Business Finland Industrial Data Excellence (InDEx) project, the Digital, Internet, Materials, and Engineering Co-Creation (DIMECC) Intelligent Industrial Data Program, and the NSF IUCRC CBL Program under Project AMaLIA, funded by Business Finland Co-Research.
\appendices
\renewcommand\thetable{A.\arabic{table}}   
\setcounter{table}{0}
\section{The proposed model variants} \label{sec:model_variants}
Table \ref{table:proposed_cnn_alternatives} summarizes the proposed CNN model variants' architecture for implementation. The variants were generated by altering the number of blocks, input image sizes, and input image color state.
\begin{table*}[!t]
\centering
\footnotesize

\caption{The proposed CNN model variants' structure. $N$C[$k\times k$] denotes a convolutional layer with $N$ kernels each of size $k\times k$, MP[$k\times k$] is Max Pooling with a $k\times k$ kernel, D$[k]$ is dropout with rate $k$, and FC[$N$] is a fully connected layer with $N$ nodes. The selected model variant (CNN9) is highlighted in bold.}
\begin{adjustbox}{max width=\textwidth}
\begin{tabu}{ccccccc}
\toprule
\textbf{Variant \#} & \textbf{Input Image Size} & \textbf{Preliminary} & \textbf{First Block} & \textbf{Second Block} & \textbf{Third Block} & \textbf{Fully Connected}
\\\midrule
CNN1 & 16x16x1 & 64C[3x3] & 64C[3x3]-MP[2x2]-D[0.4] & - & - & Flatten-D[0.4]-FC[20]-D[0.2]-Output[2]
\\\midrule
CNN2 & 16x16x1 & 64C[3x3] & 64C[3x3]-MP[2x2]-D[0.4] & 64C[3x3]-MP[2x2]-D[0.4] & - & Flatten-D[0.4]-FC[20]-D[0.2]-Output[2]
\\\midrule
CNN3 & 16x16x3 & 64C[3x3] & 64C[3x3]-MP[2x2]-D[0.4] & - & - & Flatten-D[0.4]-FC[20]-D[0.2]-Output[2]
\\\midrule
CNN4 & 16x16x3 & 64C[3x3] & 64C[3x3]-MP[2x2]-D[0.4] & 64C[3x3]-MP[2x2]-D[0.4] & - & Flatten-D[0.4]-FC[20]-D[0.2]-Output[2]
\\\midrule
CNN5 & 32x32x1 & 64C[3x3] & 64C[3x3]-MP[2x2]-D[0.4] & - & - & Flatten-D[0.4]-FC[20]-D[0.2]-Output[2]
\\\midrule
CNN6 & 32x32x1 & 64C[3x3] & 64C[3x3]-MP[2x2]-D[0.4] & 64C[3x3]-MP[2x2]-D[0.4] & - & Flatten-D[0.4]-FC[20]-D[0.2]-Output[2]
\\\midrule
CNN7 & 32x32x1 & 64C[3x3] & 64C[3x3]-MP[2x2]-D[0.4] & 64C[3x3]-MP[2x2]-D[0.4] & 64C[3x3]-MP[2x2]-D[0.4] & Flatten-D[0.4]-FC[20]-D[0.2]-Output[2]
\\\midrule
CNN8 & 32x32x3 & 64C[3x3] & 64C[3x3]-MP[2x2]-D[0.4] & - & - & Flatten-D[0.4]-FC[20]-D[0.2]-Output[2]
\\\midrule
\textbf{CNN9} & \textbf{32x32x3} & \textbf{64C[3x3]} & \textbf{64C[3x3]-MP[2x2]-D[0.4]} & \textbf{64C[3x3]-MP[2x2]-D[0.4]} & \textbf{-} & \textbf{Flatten-D[0.4]-FC[20]-D[0.2]-Output[2]}
\\\midrule
CNN10 & 32x32x3 & 64C[3x3] & 64C[3x3]-MP[2x2]-D[0.4] & 64C[3x3]-MP[2x2]-D[0.4] & 64C[3x3]-MP[2x2]-D[0.4] & Flatten-D[0.4]-FC[20]-D[0.2]-Output[2]
\\\midrule
CNN11 & 64x64x1 & 64C[3x3] & 64C[3x3]-MP[2x2]-D[0.4] & - & - & Flatten-D[0.4]-FC[20]-D[0.2]-Output[2]
\\\midrule
CNN12 & 64x64x1 & 64C[3x3] & 64C[3x3]-MP[2x2]-D[0.4] & 64C[3x3]-MP[2x2]-D[0.4] & - & Flatten-D[0.4]-FC[20]-D[0.2]-Output[2]
\\\midrule
CNN13 & 64x64x1 & 64C[3x3] & 64C[3x3]-MP[2x2]-D[0.4] & 64C[3x3]-MP[2x2]-D[0.4] & 64C[3x3]-MP[2x2]-D[0.4] & Flatten-D[0.4]-FC[20]-D[0.2]-Output[2]
\\\midrule
CNN14 & 64x64x3 & 64C[3x3] & 64C[3x3]-MP[2x2]-D[0.4] & - & - & Flatten-D[0.4]-FC[20]-D[0.2]-Output[2]
\\\midrule
CNN15 & 64x64x3 & 64C[3x3] & 64C[3x3]-MP[2x2]-D[0.4] & 64C[3x3]-MP[2x2]-D[0.4] & - & Flatten-D[0.4]-FC[20]-D[0.2]-Output[2]
\\\midrule
CNN16 & 64x64x3 & 64C[3x3] & 64C[3x3]-MP[2x2]-D[0.4] & 64C[3x3]-MP[2x2]-D[0.4] & 64C[3x3]-MP[2x2]-D[0.4] & Flatten-D[0.4]-FC[20]-D[0.2]-Output[2]
\\\bottomrule
\end{tabu}
\end{adjustbox}
\label{table:proposed_cnn_alternatives}
\end{table*}
\section{The Zhou et al. (2019) and (2021) models} \label{sec:baseline_model_1}
The proposed damage detection solution was compared to the Zhou \emph{et al.} (2019) \cite{Zhou2019} and Zhou \emph{et al.} (2021)\footnote{The adopted Zhou \emph{et al.} (2021) model is named WRIPDCNN1 in \cite{Zhou2021}.} \cite{Zhou2021} models in terms of performance and computational requirements.
These models were originally designed to detect surface damage in steel wire rope images with high performance. Although the rope material differs from our experiments (steel v.s. fiber), these detection models are still suitable for adoption due to the similarity between the two problems; they both deal with detecting damaged yarns or strands in rope images.
The original Zhou \emph{et al.} (2019) and (2021) architectures accept grayscale input images of size $64\times64$ and $96\times96$, and produce outputs of size 3 and 2, respectively.
In this work, we changed the Zhou \emph{et al.} (2019) output shape to 2 to match our problem definition, and we added six dropout layers ($0.5$ rate) to mitigate overfitting. Moreover, we increased the Zhou \emph{et al.} (2021) original two dropout rates to 0.5 and added three more dropout layers to avoid overfitting. We also created a further optimized version the models by training them with different input sizes. These results are shown in Table \ref{table:perf_results_different_structures_competing_methods}.
\section{The Schuler et al. (2022) model} \label{sec:baseline_model_2}
The proposed damage detection solution was also compared to the Schuler \emph{et al.} (2022)\footnote{The adopted Schuler \emph{et al.} (2022) model is named kDenseNet-BC L100 12ch in \cite{Schuler_2022}.} \cite{Schuler_2022} model; a highly efficient CNN-based classifier with four blocks.
The Schuler \emph{et al.} (2022) model was designed cleverly to reduce the number of required parameters while maintaining high performance; the accuracy drop was 2\% for a 55\% reduction in parameters when tested on the CIFAR-10 dataset \cite{Schuler_2022}. Therefore, we opted for this architecture for comparison because it is aligned with our design requirements; high efficiency and performance. We optimized the Schuler \emph{et al.} (2022) by trying different image input sizes. In this work, we used its original implementation. Nonetheless, we reduced its output size from 10 to 2 to match our problem definition and added dropout layers ($0.3$ rate) to minimize overfitting. Table \ref{table:perf_results_different_structures_competing_methods} shows the architecture search results.
\begin{table}[!t]
\centering
\caption{Architecture search results for competing methods with varying input image sizes on validation data. The original image sizes from Zhou \emph{et al.} (2019) and Zhou \emph{et al.} (2021) are labeled as ‘Original,’ while our optimized image sizes are labeled as ‘Optimized.’}
\begin{adjustbox}{max width=\columnwidth}
\begin{tabu}{ccccc}
\toprule
\textbf{Model} & \textbf{Variant \#} & \textbf{Input Image Size} & \textbf{Notation} & \textbf{Accuracy}
\\\midrule
Zhou et al. (2019) & 1 & 16x16x1 & & $90.1 \pm 1.2$
\\\midrule
Zhou et al. (2019) & 2 & 16x16x3 & & $93.6 \pm 1.1$
\\\midrule
Zhou et al. (2019) & 3 & 32x32x1 & & $92.9 \pm 1.5$
\\\midrule
Zhou et al. (2019) & 4 & 32x32x3 & Optimized & $94.8 \pm 0.9$
\\\midrule
Zhou et al. (2019) & 5 & 64x64x1 & Original & $93.5 \pm 0.4$
\\\midrule
Zhou et al. (2019) & 6 & 64x64x3 & & $93.6 \pm 1.9$
\\\midrule
\\\midrule
Zhou et al. (2021) & 1 & 16x16x1 & & $92.0 \pm 0.4$
\\\midrule
Zhou et al. (2021) & 2 & 16x16x3 & & $94.5 \pm 0.3$
\\\midrule
Zhou et al. (2021) & 3 & 32x32x1 & & $94.3 \pm 0.3$
\\\midrule
Zhou et al. (2021) & 4 & 32x32x3 & & $95.6 \pm 0.3$
\\\midrule
Zhou et al. (2021) & 5 & 64x64x1 & & $95.1 \pm 0.1$
\\\midrule
Zhou et al. (2021) & 6 & 64x64x3 & Optimized & $96.0 \pm 0.3$
\\\midrule
Zhou et al. (2021) & 7 & 96x96x1 & Original & $95.4 \pm 0.3$
\\\midrule
\\\midrule
Schuler et al. (2022) & 1 & 16x16x1 & & $94.7 \pm 0.5$
\\\midrule
Schuler et al. (2022) & 2 & 16x16x3 & & $95.6 \pm 0.3$
\\\midrule
Schuler et al. (2022) & 3 & 32x32x1 & & $93.8 \pm 2.4$
\\\midrule
Schuler et al. (2022) & 4 & 32x32x3 & Optimized & $96.2 \pm 0.2$
\\\bottomrule
\end{tabu}
\end{adjustbox}
\label{table:perf_results_different_structures_competing_methods}
\end{table}
\bibliography{Sections/references}
\begin{IEEEbiography}[{\includegraphics[width=1in,height=1.25in,clip,keepaspectratio]{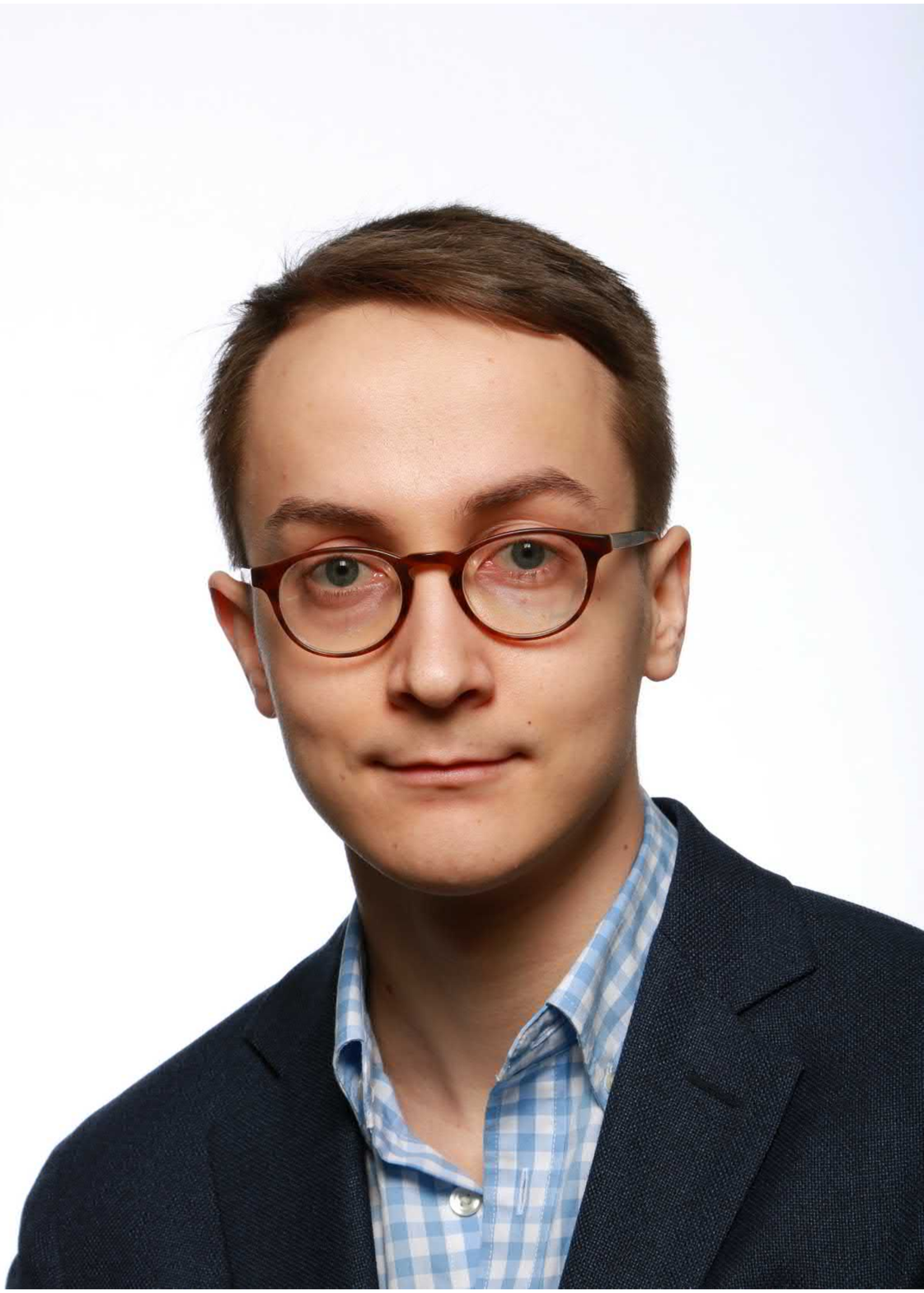}}]{Tuomas Jalonen} (Graduate Student Member, IEEE)
received the B.Sc. and M.Sc. degrees in Mechanical Engineering from Tampere University, Finland, in 2017 and 2019, respectively. He is currently pursuing a Ph.D. degree with the Faculty of Information Technology and Communication Sciences at Tampere University, specializing in machine learning. His research interests include machine learning methods, signal processing, deep learning, computer vision, time-series analysis, and their applications in industrial systems.
\end{IEEEbiography}
\begin{IEEEbiography}[{\includegraphics[width=1in,height=1.25in,clip,keepaspectratio]{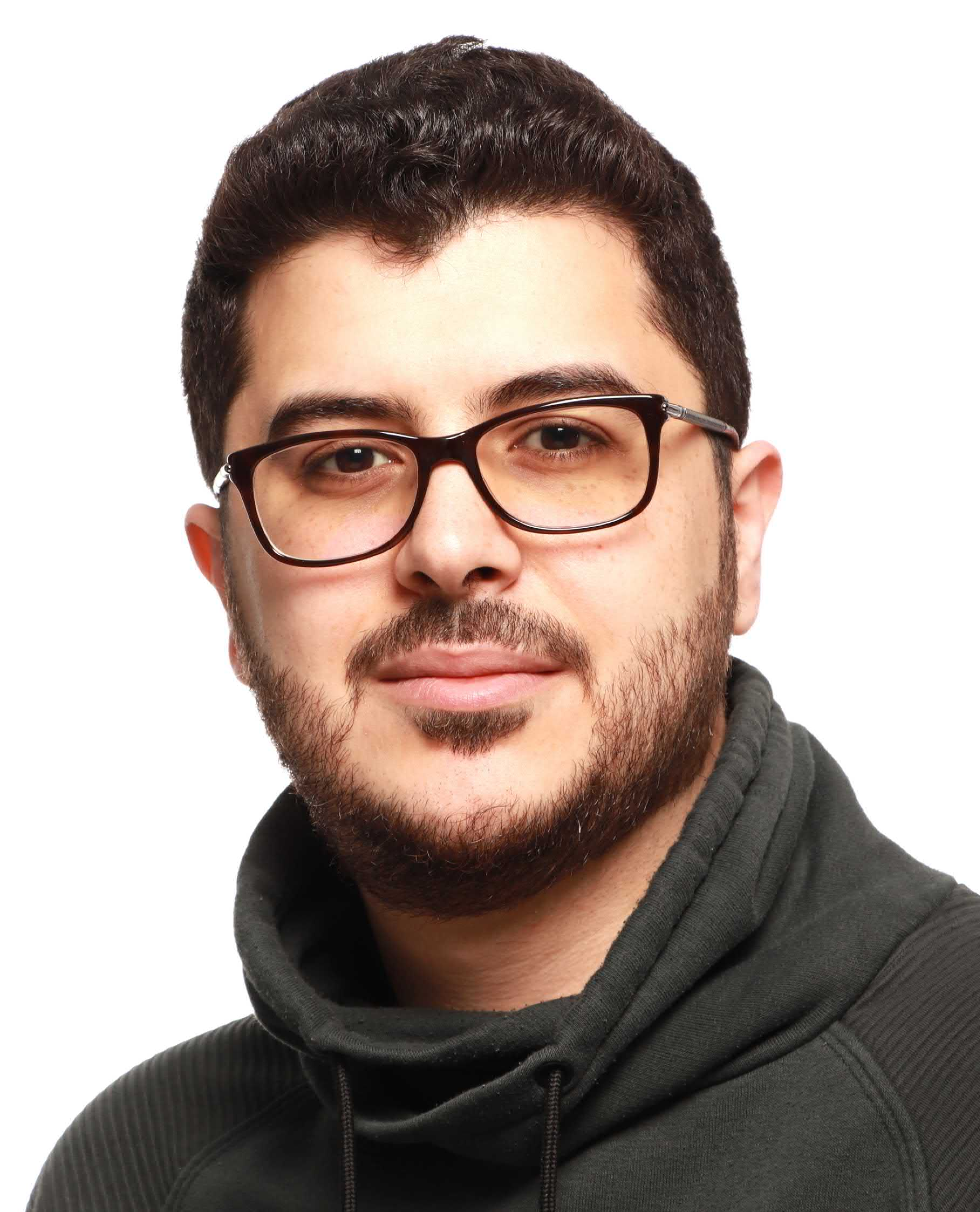}}]{Mohammad Al-Sa'd}
(Senior Member, IEEE) received his B.Sc. and M.Sc. degrees in Electrical Engineering from Qatar University, Qatar, in 2012 and 2016, respectively, and his PhD degree in Electrical Engineering and Computing Sciences from Tampere University, Finland, in 2022. He specializes in signal processing and is currently pursuing his postdoctoral fellowship at the Department of Physiology, University of Helsinki, Finland, and at the Faculty of Information Technology and Communication Sciences, Tampere University, Finland. He has served as a technical reviewer for several journals, including IEEE Transactions on Signal Processing, IEEE Transactions on Instrumentation \& Measurement, IEEE Transactions on Artificial Intelligence, Digital Signal Processing, Signal Processing, Biomedical Signal Processing and Control, and IEEE Access. His research interests include time-frequency signal theory, machine learning, neuroscience, electroencephalogram analysis and processing, information flow and theory, signal modeling, and optimization.
\end{IEEEbiography}
\begin{IEEEbiography}[{\includegraphics[width=1in,height=1.25in,clip,keepaspectratio]{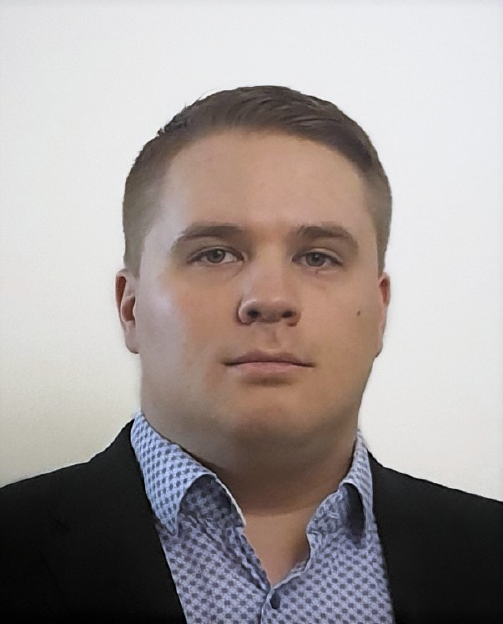}}]{Roope Mellanen} received the B.Sc. and M.Sc. degrees in automation from Tampere University of Technology, Finland, in 2016 and 2018, respectively. He is currently working as a Research Specialist for Konecranes Corporation. His research interest includes machine learning, computer vision, and signal processing and analytics.
\end{IEEEbiography}
\begin{IEEEbiography}[{\includegraphics[width=1in,height=1.25in,clip,keepaspectratio]{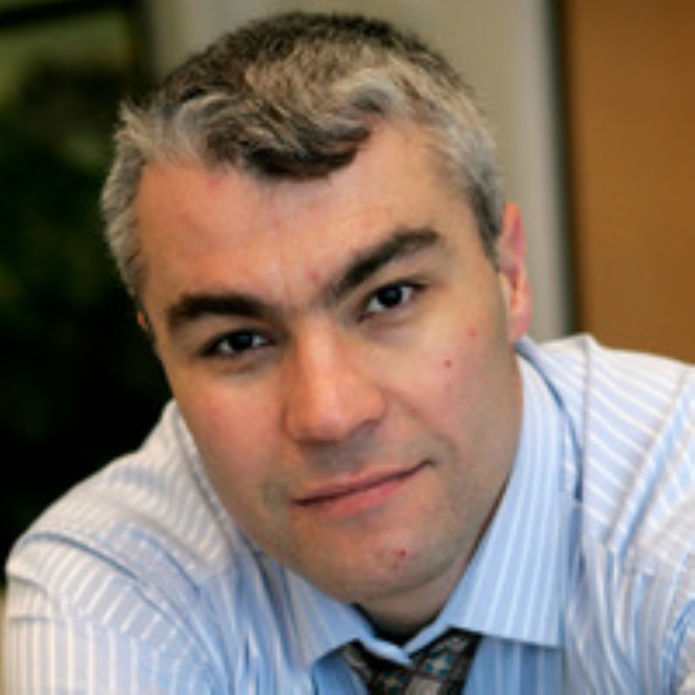}}]{Serkan Kiranyaz}
(Senior Member, IEEE) is a Professor with Qatar University, Doha, Qatar. He published two books, five book chapters, more than 80 journal articles in high impact journals, and 100 articles in international conferences. He made contributions on evolutionary optimization, machine learning, bio-signal analysis, computer vision with applications to recognition, classification, and signal processing. He has coauthored the articles which have nominated or received the “Best Paper Award” in ICIP 2013, ICPR 2014, ICIP 2015, and IEEE Transactions on Signal Processing (TSP) 2018. He had the most-popular articles in the years 2010 and 2016, and most-cited article in 2018 in IEEE Transactions on Biomedical Engineering. From 2010 to 2015, he authored the 4th most-cited article of the Neural Networks journal. His research team has won the second and first places in PhysioNet Grand Challenges 2016 and 2017, among 48 and 75 international teams, respectively. His theoretical contributions to advance the current state of the art in modeling and representation, targeting high long-term impact, while algorithmic, system level design and implementation issues target medium and long-term challenges for the next five to ten years. He in particular aims at investigating scientific questions and inventing cutting-edge solutions in “personalized biomedicine” which is in one of the most dynamic areas where science combines with technology to produce efficient signal and information processing systems.
\end{IEEEbiography}
\begin{IEEEbiography}[{\includegraphics[width=1in,height=1.25in,clip,keepaspectratio]{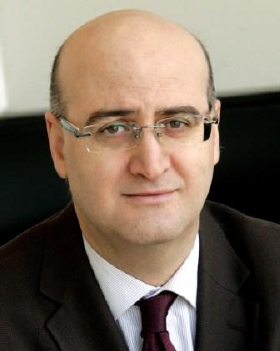}}]{Moncef Gabbouj}
(Fellow Member, IEEE) received the B.S. degree from Oklahoma State University, Stillwater, OK, USA, in 1985, and the M.S. and Ph.D. degrees from Purdue University, in 1986 and 1989, respectively, all in electrical engineering. He is a Professor of signal processing with the Department of Computing Sciences, Tampere University, Tampere, Finland. He was an Academy of Finland Professor from 2011 to 2015. His research interests include big data analytics, multimedia content-based analysis, indexing and retrieval, artificial intelligence, machine learning, pattern recognition, nonlinear signal and image processing and analysis, voice conversion, and video processing and coding. Dr. Gabbouj is a member of the Academia Europaea and the Finnish Academy of Science and Letters. He is the past Chairman of the IEEE CAS TC on DSP and the Committee Member of the IEEE Fourier Award for Signal Processing. He served as an Associate Editor and the Guest Editor of many IEEE, and international journals and a Distinguished Lecturer for the IEEE CASS. He is the Finland Site Director of the NSF IUCRC funded Center for Visual and Decision Informatics (CVDI) and leads the Artificial Intelligence Research Task Force of the Ministry of Economic Affairs and Employment funded Research Alliance on Autonomous Systems (RAAS).
\end{IEEEbiography}
\end{document}